\documentclass[10pt,twocolumn,letterpaper]{article}

\usepackage[pagenumbers]{iccv} % To force page numbers, e.g. for an arXiv version
\usepackage{multirow}
\usepackage{xcolor}
\usepackage{colortbl}
\usepackage{adjustbox}
\usepackage{pifont}
\usepackage{makecell}
\usepackage{array}
\usepackage{tikz}
\usepackage{amssymb}
\usepackage{mathabx}
\usepackage{fontawesome}

\definecolor{my_green}{RGB}{51,102,0}
\definecolor{my_red}{RGB}{204, 0, 0}
\renewcommand{\checkmark}{\textcolor{my_green}{\ding{51}}} % ✔
\newcommand{\crossmark}{\textcolor{my_red}{\ding{55}}} % ✘

\newcommand{\nonumfootnote}[1]{%
  \begingroup
  \renewcommand{\thefootnote}{}% 临时取消编号显示
  \footnote{#1}% 插入无编号脚注
  \addtocounter{footnote}{-1}% 恢复计数器（确保后续编号正确）
  \endgroup
}

\definecolor{iccvblue}{rgb}{0.21,0.49,0.74}
\usepackage[pagebackref,breaklinks,colorlinks,allcolors=iccvblue]{hyperref}

\def\paperID{806} % *** Enter the Paper ID here
\def\confName{ICCV}
\def\confYear{2025}

\title{FAVOR-Bench: A Comprehensive Benchmark for Fine-Grained \\ Video Motion Understanding}
\author{
Chongjun Tu$^{1\dagger}$ \quad Lin Zhang$^{1, 3\dagger}$ \quad Pengtao Chen$^{1\dagger}$ \quad Peng Ye$^2$ \quad Xianfang Zeng$^{3 \text{\ding{168}}}$ \\
Wei Cheng$^3$ \quad Gang Yu$^3$ \quad Tao Chen$^{1*}$ \\
\vspace{2mm}
$^{1}$ Fudan University \quad $^{2}$ The Chinese University of Hong Kong \quad $^{3}$ StepFun
}

\begin{document}
\maketitle
\begin{abstract}

Multimodal Large Language Models (MLLMs) have shown remarkable capabilities in video content understanding but still struggle with fine-grained motion comprehension.
To comprehensively assess the motion understanding ability of existing MLLMs, we introduce FAVOR-Bench, comprising 1,776 videos with structured manual annotations of various motions.
Our benchmark includes both close-ended and open-ended tasks. For close-ended evaluation, we carefully design 8,184 multiple-choice question-answer pairs spanning six distinct sub-tasks. 
For open-ended evaluation, we develop both a novel cost-efficient LLM-free and a GPT-assisted caption assessment method, where the former can enhance benchmarking interpretability and reproducibility.
Comprehensive experiments with 21 state-of-the-art MLLMs reveal significant limitations in their ability to comprehend and describe detailed temporal dynamics in video motions. 
To alleviate this limitation, we further build FAVOR-Train, a dataset consisting of 17,152 videos with fine-grained motion annotations. 
The results of finetuning Qwen2.5-VL on FAVOR-Train yield consistent improvements on motion-related tasks of TVBench, MotionBench and our FAVOR-Bench.
Comprehensive assessment results demonstrate that the proposed FAVOR-Bench and FAVOR-Train provide valuable tools to the community for developing more powerful video understanding models. 
Project page: \href{https://favor-bench.github.io/}{https://favor-bench.github.io/}.

\end{abstract}

\nonumfootnote{
\hspace{-6mm} $^{\text{\ding{168}}}$Xianfang Zeng is the project leader. \\
$^{\dagger}$Equal contribution. $^*$Corresponding author. \\
Work was done when interned at StepFun.}

\section{Introduction}
\label{sec:intro}

\begin{figure}[t]
    \centering
    \includegraphics[width=\linewidth]{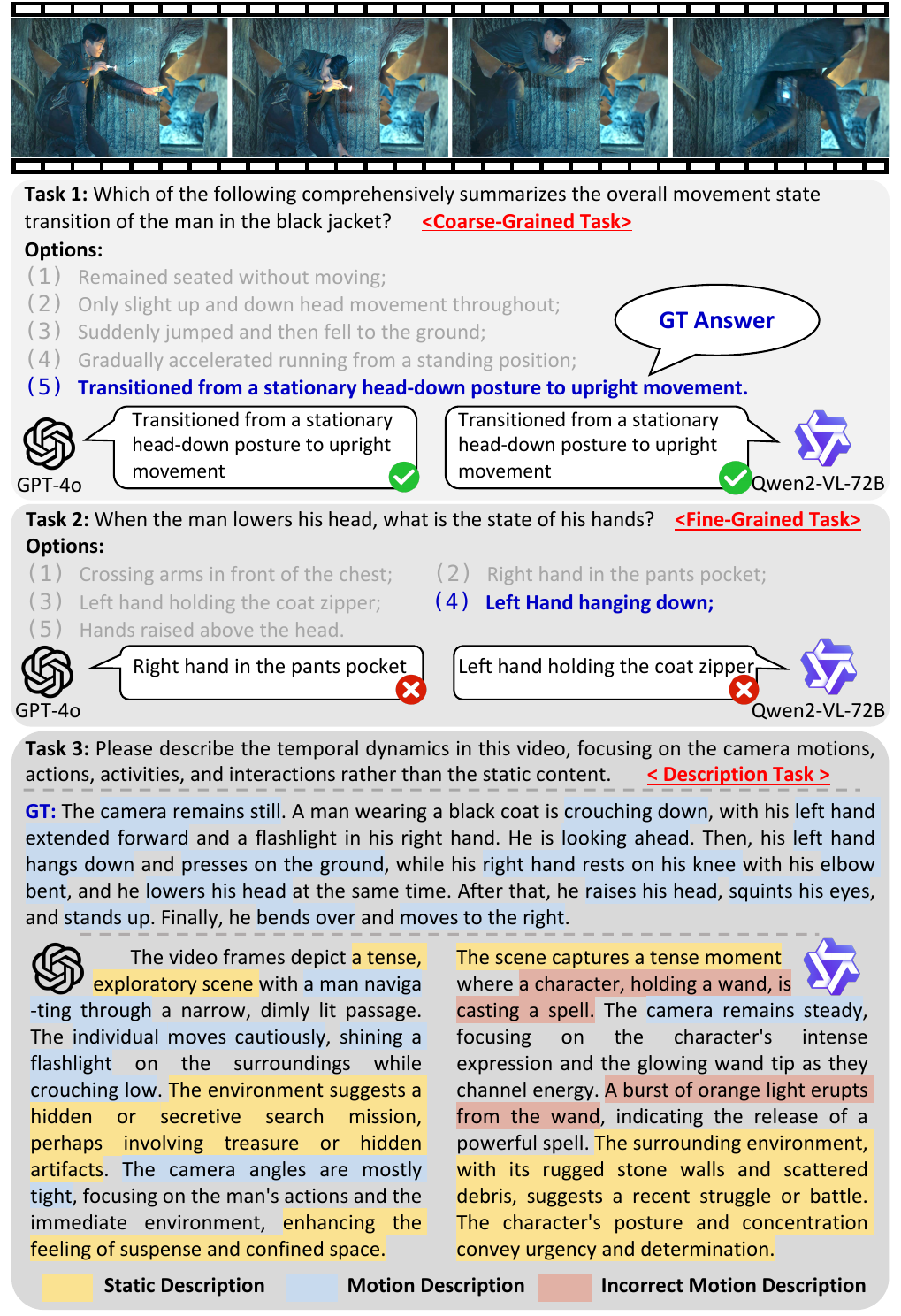}
    \vspace{-6mm}
    \caption{Illustration of motion understanding capabilities of proprietary and open-source MLLMs. Both models correctly answer the coarse-grained summarization question (Task 1), but fail to resolve fine-grained action detail question (Task 2). 
    For the open-ended description task (Task 3), despite required to focus on temporal dynamics, the responses emphasize static content, and the motion descriptions are either coarse-grained or contain errors.}
    \label{fig:motivation}
    \vspace{-5mm}
\end{figure}

Multimodal Large Language Models (MLLMs) have demonstrated remarkable video understanding capabilities~\cite{chen2024internvl2.5,zhang2025videollama3,yang2024qwen2.5}.
The emergence of high-quality video-text datasets~\cite{nan2024openvid_dataset,chen2024panda_dataset,wang2024koala_dataset} has further promoted their development and powered various downstream applications like motion recognition, caption generation and video generation~\cite{jia2022egotaskqa,yin2023lamm_task,huang2024vbench_task,li2024videovista_task,zhou2024streaming_task,athar2024vicas_task}.
To effectively evaluate the capabilities of these models, kinds of benchmarks with different focuses have been developed, such as comprehensive capabilities~\cite{fu2024video-mme,li2024mvbench}, long-video understanding~\cite{zhou2024mlvu,grauman2024ego}, and video reasoning~\cite{wu2024star,xie2024funqa}.

Despite these advances in video understanding benchmarks, the evaluation of fine-grained video motion understanding remains under-explored, which is a critical capability for fields that require precise understanding and control (such as embodied imitation learning~\cite{wang2023mimicplay,cui2023play_mimic} and text-image to video (TI2V)~\cite{hu2022make_ti2v,ni2024ti2v}).
As shown in~\cref{fig:motivation}, MLLMs are capable of identifying the overall behavior of the subjects in a video, but they struggle with problems related to fine-grained motions. 
For open-ended description tasks, even when explicitly instructed to focus on the temporal dynamics in the video, models predominantly emphasize static content and often lack fine-grained analysis of the motions and activities.
Traditional datasets like ActivityNet-QA~\cite{yu2019activitynet-qa} and motion-related subsets in current  benchmarks~\cite{li2024mvbench,zhou2024mlvu} primarily focus on the event-level granularity.
The concurrent work MotionBench~\cite{hong2025motionbench} proposes to evaluate motion-level perception through multiple-choice questions and focus on third-person perspective videos.
However, MotionBench lacks the evaluation of ego-centric videos and open-ended caption tasks.

To mitigate the above concerns, we propose FAVOR-Bench, a comprehensive benchmark for \textbf{f}ine-gr\textbf{a}ined \textbf{v}ideo m\textbf{o}tion unde\textbf{r}standing.
FAVOR-Bench includes 1,776 videos covering different types, including daily-life records, ego-centric videos, TV series, and animations. It comprises both close-ended and open-ended tasks.
For the close-ended evaluation, FAVOR-Bench has carefully curated 8,184 challenging multiple-choice question-answer (QA) pairs spanning six distinct tasks, which are obtained in a semi-automatic pipeline.
Specifically, we employ the powerful DeepSeek-R1~\cite{guo2025deepseek} to generate initial QA pairs.
Then, we propose to perform blind filtering and single-frame filtering to remove simple questions that can be solved by common sense or with a few specific frames (see~\cref{sec:qa generation} for the details).
Finally, we conduct strict manual verification to ensure the quality of QA pairs.
For open-ended evaluation, we construct fine-grained manual annotations for each video.
In addition to commonly adopted GPT-assisted evaluation~\cite{zhou2024mlvu,sun2024aligning_gpt_assist}, we propose a novel LLM-free evaluation framework, which aims to reduce the evaluation cost of generative tasks and enhance evaluation interpretability and reproducibility.

Comprehensive results on FAVOR-Bench reveal notable limitations in current video understanding models' fine-grained motion understanding capabilities.
On close-ended tasks, 
Gemini-1.5 achieves the highest accuracy (49.87\%), while Qwen2.5-VL-72B performs best (48.14\%) among the open-source models. 
% 下面这句可能需要改进，不太会写。performance is unsatisfactory / low 
For open-ended evaluation, the performance of evaluated models is also below the practical deployment expectations for real-world applications, i.e., lower than 5 out of 10 on the GPT-assisted evaluation, and below 60\% on our LLM-free evaluation.
To bridge this gap and promote the development of fine-grained motion comprehension, we further build FAVOR-Train, a dataset that contains 17,152 videos and fine-grained manually annotated captions.
By performing supervised fine-tuning (SFT) on Qwen2.5-VL~\cite{yang2024qwen2.5} with FAVOR-Train, we improve the close-ended performance by $1.37\%$ and open-ended performance by $0.2$ (GPT-assisted evaluation) and $7.9\%$ (LLM-Free evaluation) on FAVOR-Bench compared to the Qwen2.5-VL baseline. 
We further evaluate the fine-tuned model on TVBench~\cite{cores2024tvbench} and MotionBench~\cite{hong2025motionbench}.
Notably, we achieve consistent performance improvement on motion-related tasks.
These evaluations demonstrate that FAVOR-Train can effectively strengthen fine-grained motion comprehension, showcasing its value for developing more powerful video understanding models.

Our contribution can be concluded from three aspects:
\begin{itemize}
    \item We present FAVOR-Bench, the first fine-grained video motion understanding benchmark with both close-ended and open-ended evaluation. It comprises 1,776 carefully curated videos with fine-grained annotations, based on which 8,184 multi-choice QA pairs across 6 dimensions are constructed as close-ended tasks. 
    For open-ended evaluation, aside from GPT-assisted evaluation, we pioneer a novel cost-efficient LLM-free evaluation framework, which provides more interpretable evaluation and serves as an important supplement for generative tasks in video motion understanding. 
    \item Comprehensive evaluation on the proposed FAVOR-Bench exposes critical limitations in the fine-grained motion understanding capabilities of current MLLMs. Even the most powerful models achieve less than 50\% accuracy on close-ended tasks and show notable weaknesses in open-ended descriptions, highlighting substantial room for further development.
    \item We further construct FAVOR-Train, a training dataset with 17,152 videos and manual captions specifically designed for fine-grained video motion understanding. Fine-tuning Qwen2.5-VL with FAVOR-Train can lead to consistent performance improvements on our proposed FAVOR-Bench and motion-related tasks of existing benchmarks, demonstrating the effectiveness of this dataset. It can promote the community to develop stronger video understanding models.
\end{itemize}

\section{Related Work}
\label{sec:related}

\begin{figure*}
    \centering
    \includegraphics[width=\linewidth]{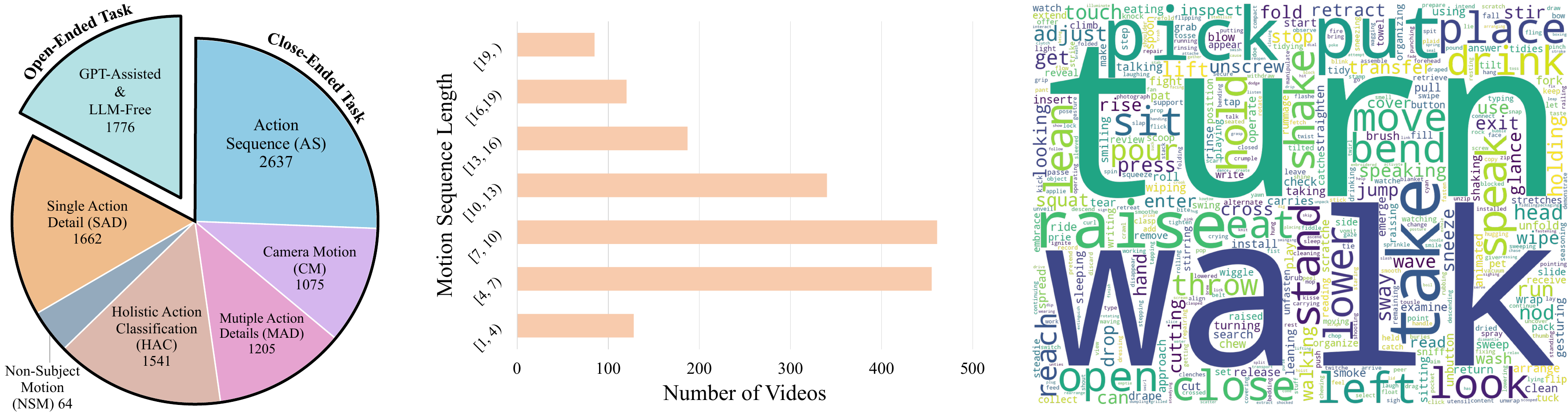}
    \vspace{-6mm}
    \caption{Data statistics of FAVOR-Bench. 
    \textbf{Left:} Task type distribution across close-ended and open-ended evaluation in FAVOR-Bench. 
    \textbf{Middle:} Distribution of motion sequence length per video. 
    \textbf{Right:} The word cloud statistics of motion vocabularies in FAVOR-Bench.}
    \label{fig:dataset overview}
    \vspace{-3mm}
\end{figure*}

\subsection{MLLMs for Video Understanding}

Recent advancements in Multimodal Large Language Models (MLLMs) have enhanced video understanding capabilities through module designs and scalable training paradigms.
Models include VideoLLaMA3~\cite{cheng2024videollama2,zhang2025videollama3} pioneer vision-centric designs, dynamically adapting tokenization strategies to capture fine-grained spatial details and temporal dynamics. 
The Tarsier series~\cite{wang2024tarsier,yuan2025tarsier2} shows that combining CLIP encoders with LLMs, along with temporal alignment techniques, enables precise video description and causal reasoning.
Qwen2-VL~\cite{wang2024qwen2} and InternVL 2.5~\cite{chen2024internvl2.5} unify text, image, and video processing through dynamic resolution mechanisms and propose advanced positional embeddings, which supports high-resolution inputs and better multimodal integration. 
Besides, the performance across various video tasks of MLLMs also benefits from larger-scale of training data and parameters~\cite{yang2024qwen2.5, chen2024internvl1.5}.
These rapid advancements underscore the necessity for more challenging benchmarks to evaluate specific aspects of MLLM's video understanding capabilities.

\subsection{Video Understanding Benchmarks}

With the enhancement of models' video understanding capabilities, various benchmarks have been developed to evaluate their performance.
% from different aspects.
Traditional benchmarks primarily evaluate basic video understanding and short caption generation capabilities~\cite{xu2016msr-vtt,caba2015activitynet}.
Subsequent works reveal the single frame bias~\cite{huang2018makes_single_frame,lei2023revealing_single_frame} and emphasize the importance of temporal dynamics for evaluations~\cite{sevilla2021only_time_can_tell}.
Recently, benchmarks focusing on different aspects have been constructed, as depicted in~\cref{tab:comparison}. 
Comprehensive benchmarks like MVBench~\cite{li2024mvbench} and Video-MME~\cite{fu2024video-mme} evaluate the general video understanding capabilities of models. 
Besides, benchmarks focused on certain specific challenging scenarios have been constructed, such as long-form video understanding~\cite{zhou2024mlvu,mangalam2023egoschema}, counter-intuitive reasoning~\cite{xie2024funqa}, and ego-centric video understanding~\cite{mangalam2023egoschema}.

For motion understanding in videos, ActivityNet-QA~\cite{caba2015activitynet} introduces VideoQA with manually annotated datasets. 
NExT-QA~\cite{xiao2021next-qa} further evaluates the causal and temporal reasoning abilities of QA models.
Besides, some comprehensive benchmarks include subsets for action and motion understanding~\cite{zhou2024mlvu,li2024mvbench,cores2024tvbench}
However, these works primarily evaluate the models at the event level, lacking consideration for more fine-grained motions. 
The concurrent work MotionBench~\cite{hong2025motionbench} tries to bridge this gap through multiple-choice evaluation of motion-level perception.
Compared to MotionBench, our proposed FAVOR-Bench not only expands scenario diversity with ego-centric videos but also includes both close-ended (e.g., multiple-choice questions) and open-ended (video description) tasks, thereby providing a more comprehensive evaluation of models' fine-grained motion understanding and description abilities.

\section{FAVOR-Bench: Fine-Grained Video Motion Understanding Benchmark}
\label{sec:method}
This section presents FAVOR-Bench, a comprehensive benchmark for fine-grained video motion understanding. 
We start with a brief overview of FAVOR-Bench. 
Then, we provide detailed descriptions of the dataset curation and evaluation tasks.

\subsection{Overview}

\begin{table*}[t!]
\centering
\resizebox{0.99\linewidth}{!}
{
\vspace{5pt}
\renewcommand{\arraystretch}{1.0} % Slightly reduced row height
\begin{tabular}{lcccccccc}
\toprule
\raisebox{-3pt}{\multirow{2}{*}{\textbf{Benchmarks}}} & \raisebox{-3pt}{\multirow{2}{*}{\textbf{\#Videos}}} & \multicolumn{3}{c}{\textbf{Video Type}} & \raisebox{-3pt}{\multirow{2}{*}{\textbf{Fine-Grained Motion}}} & \raisebox{-3pt}{\multirow{2}{*}{\textbf{\#Close-Ended QA}}} & \multicolumn{2}{c}{\makecell{\textbf{Open-Ended Evaluation}}}  \\
\cmidrule{3-5} \cmidrule{8-9}
 & & \textbf{Third-Person} & \textbf{Ego-Centric} & \textbf{Simulation} & & & \textbf{GPT-Assisted} & \textbf{LLM-Free} \\

\midrule
MVBench~\cite{li2024mvbench} & 4,000 & \checkmark & \crossmark & \checkmark & \crossmark & 4,000 & \crossmark & \crossmark  \\
TVBench~\cite{cores2024tvbench} & 2,525 & \checkmark & \crossmark & \checkmark & \crossmark & 2,525 & \crossmark & \crossmark  \\
EgoSchema~\cite{mangalam2023egoschema} & 5,031 & \crossmark & \checkmark & \crossmark & \crossmark & 5,031 & \crossmark & \crossmark  \\
EgoTaskQA~\cite{jia2022egotaskqa} & 2,315 & \crossmark & \checkmark & \crossmark & \checkmark & 40,322 & \crossmark & \crossmark  \\
MLVU~\cite{zhou2024mlvu} & 1,730 & \checkmark & \crossmark & \checkmark & \crossmark & 3,102 & \checkmark & \crossmark  \\
MovieChat-1K~\cite{song2024moviechat} & 130 & \checkmark & \crossmark & \crossmark & \crossmark & 1,950 & \crossmark & \crossmark  \\
MotionBench~\cite{hong2025motionbench} & 5,385 & \checkmark & \crossmark & \checkmark & \checkmark & 8,052 & \crossmark & \crossmark  \\

\midrule
\textbf{FAVOR-Bench} & 1,776 & \checkmark & \checkmark & \checkmark & \checkmark & 8,184 & \checkmark & \checkmark  \\
\bottomrule
\end{tabular}
}
\vspace{-4pt}
\caption{Comparison of FAVOR-Bench with existing video understanding benchmarks. \textbf{\#Videos} and \textbf{\#Close-Ended QA} refer to the number of videos and close-ended question-answer pairs respectively.
FAVOR-Bench covers wide video types (Third-Person, Ego-Centric, Simulation) while focusing on fine-grained motion understanding. 
Moreover, FAVOR-Bench provides comprehensive evaluation, including close-ended QA and open-ended tasks (both GPT-assisted evaluation and our novel LLM-Free framework).
}
\vspace{-8pt}
\label{tab:comparison}
\end{table*}

% The benchmark introduces two evaluation protocols: 1) open-ended assessment with our novel LLM-free metric, and 2) standardized GPT-4 based scoring. This dual-assessment framework enables comprehensive evaluation of both motion understanding precision and descriptive capabilities.

% As shown in \Cref{fig:dataset overview}, our dataset exhibits three key characteristics: (1) balanced task distribution across difficulty levels, (2) prevalence of multi-motion video instances (3-5 concurrent motions in 68% videos), and (3) rich semantic coverage of 217 fine-grained motion categories.

FAVOR-Bench consists of 1,776 carefully curated videos spanning diverse domains. 
The length of these videos sums up to 10.2 hours, with an average duration of 20.6 seconds. 
Through a semi-automatic pipeline, we construct 8,184 challenging QA pairs with fine-grained structured manual annotations, systematically challenging models via six motion-centric tasks.
% Based on these videos, we first conduct fine-grained manual annotations and then generate 8,184 QA pairs in a semi-automatic pipeline to challenge the models through six tasks related to motion understanding. 
In addition, FAVOR-Bench includes open-ended evaluation, including the widely adopted GPT evaluation and our novel LLM-free evaluation framework.
% for which we design an novel LLM-free metric. Combined with the widely adopted GPT evaluation, 
Through these tasks, we comprehensively assess the fine-grained motion understanding and description capabilities of models.
\Cref{fig:dataset overview} shows the data statistics of FAVOR-Bench, including the task type distribution, the distribution of motion counts per video, and the word cloud statistics of motion vocabularies.

\subsection{Dataset Curation}
\label{sec:dataset curation}
In this section, we elaborate on the dataset curation process of FAVOR-Bench, including data collection, data filtering, and manual annotation.

\noindent \textbf{Data Collection and Filtering.}
% 写数据采集或者过滤的时候可以提到一下：为了确保细粒度的动作理解和标注质量，我们倾向于选择动作丰富且时长不太长的视频。
% 不同种类的视频中动作分布不一样(还包括单人、多人数量不一样)，ego-centric视频进一步带来了额外的挑战。（可参考motionbench）
The raw videos we collected consist of four types: daily-life records, TV series, animations and ego-centric videos. 
To ensure fine-grained motion understanding and annotation quality, we choose videos with rich motions and relatively short durations. 
For daily-life record, 868 videos are sampled from Charades~\cite{sigurdsson2016hollywood_charades} with the highest quality scores (quality scores are offered by Charades). 
For TV series and animations, video clips with high motion quality are acquired by our own with a comprehensive pipeline including scene-aware cropping, optical flow-based filtering, and manual curation. 574 clips from TV-series and 138 clips from animations are selected. 
For Egocentric videos, we select EgoTaskQA~\cite{jia2022egotaskqa} as the data source and randomly sample 196 videos.
More details of video curation and filtering are provided in the supplementary material.

\noindent \textbf{Manual Annotation.}
% 可以提一下雇佣了多少高学历的本科生（专科生），全职还是兼职，用了多长时间标注
We hired 8 highly educated personnel for a two-week full-time labeling process. 
Manual inspection-revision mechanism has been introduced to ensure the quality of the annotations. 
For each video, the structured annotation contains the following components: 
1) Subjects (such as man, woman, dog, first-person subject, etc.) involved in motion or action, with up to three attributes (like wearing, color, etc.) 
2) The motion list of each subject and the start \& end time of each motion, specified in seconds. 
3) Camera motions and their start \& end time, specified in seconds. 
4) The overall caption of the video, which include all the annotated subjects, motions and camera motions mentioned above.

\subsection{Close-Ended Evaluation}
\label{sec:close_eval}

\begin{figure*}
    \centering
    \includegraphics[width=0.93\linewidth]{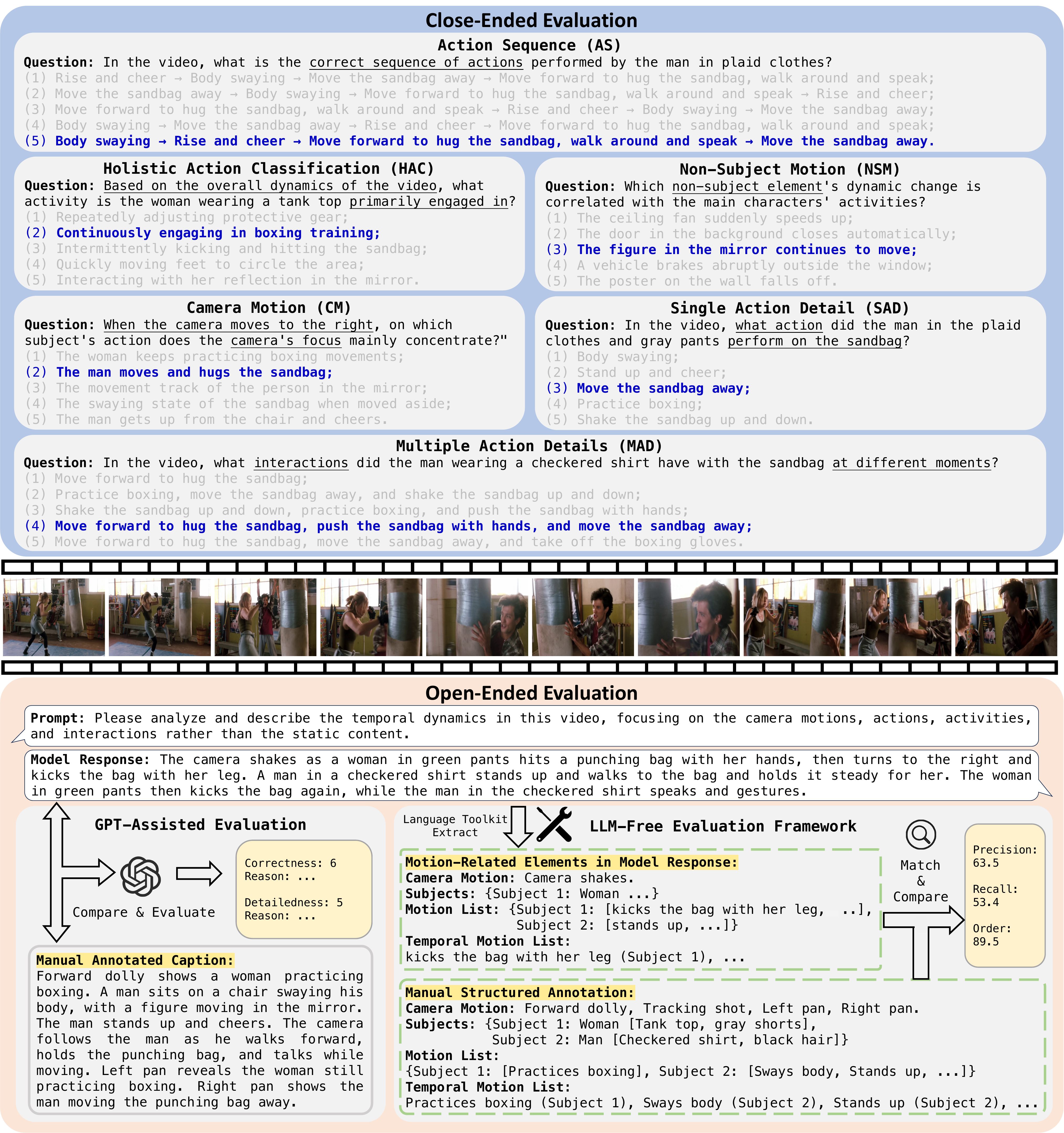}
    \caption{Overview of evaluation tasks. FAVOR-Bench comprises close-ended and open-ended evaluations.
    The close-ended evaluation is composed of six tasks, focusing on different aspects of fine-grained motion understanding.
    The open-ended evaluation comprises a GPT-assisted evaluation and a novel LLM-free framework. In the GPT-assisted evaluation, model responses are directly compared with manual captions. The LLM-free framework parses structured motion elements from responses and compares them with the structured annotations.}
    \label{fig:task definition}
    \vspace{-3mm}
\end{figure*}

\subsubsection{Task Definition}
\label{sec:close_task_def}
Close-ended tasks are widely adopted in existing benchmarks to evaluate specific capabilities.
FAVOR-Bench proposes to examine six critical dimensions of fine-grained motion understanding through carefully designed tasks, which are formatted as multiple-choice questions (illustrated in the upper part of~\cref{fig:task definition}). 
The performance is then measured by the answer accuracy. 
The detailed explanations of each task are as follows:

\noindent \textbf{Action Sequence (AS).} 
The action sequence task focuses on understanding the temporal dynamics in the video. 
In this task, one or more subjects in the video may perform a series of complex actions, and the models are required to compare which action occurs first or answer the complete action sequence of a certain subject in the video.

\noindent \textbf{Holistic Action Classification (HAC).}
The holistic action classification task requires models to answer the core action of the subjects in the video. This task is similar to traditional video action classification and recognition tasks, focusing on the global action summarization ability.

\noindent \textbf{Single Action Detail (SAD).}
This task examines the moment-specific detail recognition ability. The model will be asked about the state of the subjects at a specific moment, as well as the interaction between the subject and an object.

\noindent \textbf{Multiple Action Details (MAD).}
This task focuses on evaluating the ability to compare and analyze details across multiple moments. 
The model will be required to answer the changes in the actions and states of the subject over time, or the interactions between the subject and multiple objects.

\noindent \textbf{Camera Motion (CM).}
The camera motion task examines the understanding of viewpoint dynamics, focus shifts, and their coordination with subject actions in the video.
This capability is equally important for fine-grained action understanding, as camera motion may affect the visibility of the subjects or cause a subject switch.

\noindent \textbf{Non-Subject Motion (NSM).}
This task focuses on evaluating the environmental context awareness of models, such as the movements and behaviors of non-subject elements (including background objects and passersby) in the video.
The non-subject motions can serve as peripheral cues to refine the primary motion understanding, especially for real-world scenarios.

\subsubsection{QA Generation}
\label{sec:qa generation}
Based on the curated video dataset and fine-grained structured annotations, we adopt a three-stage semi-automatic pipeline to construct QA pairs: 1) Automatic Question-Answer Generation, 2) Blind Filtering and Single-Frame Filtering, and 3) Manual Verification. 
After the above steps, FAVOR-Bench constructs an average of 4.6 multiple-choice questions for each video.

\noindent \textbf{Automatic Question-Answer Generation.}
For each of the six tasks, we design distinct prompt templates to generate QA pairs from the annotation metadata using DeepSeek-R1~\cite{guo2025deepseek}.
Our guidelines for automatic QA generation include two critical requirements: 1) maximizing question diversity to cover more video content, and 2) crafting challenging distractors without compromising the uniqueness of the correct answers.
The specific prompt templates can be found in the supplementary material.
Through this process, we obtain 20,402 multiple-choice QA pairs in total.

% \vspace{-2mm} 
\noindent \textbf{Blind Filtering and Single-Frame Filtering.}
To ensure the benchmark's quality and challenge, we design a two-stage filtering framework comprising blind filtering and single-frame filtering to help remove low-quality QA pairs.
Specifically, our analysis reveals that a subset of the generated questions can be resolved solely through common sense and language priors, without any visual information.
Furthermore, the single-frame bias~\cite{huang2018makes_single_frame,lei2023revealing_single_frame}, which refers to scenarios where a single frame suffices to answer the video-understanding question, can also impact motion understanding evaluation in videos.
For blind filtering, we choose Qwen2-72B~\cite{yang2024qwen2technicalreport} as the representative LLM to answer the questions without visual input and remove the correctly addressed questions.
Subsequently, for single-frame filtering, the remaining QA pairs are fed into GPT-4o~\cite{hurst2024gpt4o} together with 5 frames that are uniformly sampled from the corresponding video. The correctly answered questions are further filtered out.
While this filtering framework might inadvertently exclude instances where the answers are correct by random chance, it effectively enhances the quality of the constructed close-ended tasks.
After this process, 12,096 QA pairs remain.

\noindent \textbf{Manual Verification.}
In this stage, annotators verify all QA pairs from three perspectives: question-video relevance, answer correctness, and answer uniqueness. 
Ambiguous or incorrect QA pairs are discarded.
We further calibrate option distributions to mitigate position and frequency biases. 
FAVOR-Bench ultimately comprises 8,184 multiple-choice questions.

\subsection{Open-Ended Evaluation}
Beyond close-ended tasks that provide constrained options, FAVOR-Bench further challenges the models' comprehensive fine-grained motion understanding and description capabilities through generative tasks.
Our open-ended evaluation includes two types of metrics: GPT-assisted evaluation and a novel LLM-free evaluation (illustrated in the lower part of~\cref{fig:task definition}).

\subsubsection{GPT-Assisted Evaluation}
Following benchmarks with generative tasks~\cite{sun2024aligning_gpt_assist,zhou2024mlvu}, we employ GPT-4o~\cite{hurst2024gpt4o} to assist in the open-ended evaluation.
Specifically, we prompt the model being evaluated to describe the temporal dynamics in the video, including subject actions, camera motions, etc.
Then, we input the generated responses and the manually crafted descriptions into GPT-4o for comparison and scoring from both correctness and detailedness perspectives.
To reduce randomness and enhance the evaluation's robustness, we set the scoring range (from 1 to 10) in the prompt template and define the specific criteria for each score level.
The prompt template is provided in the supplementary material.

\subsubsection{LLM-Free Evaluation}
While using the powerful GPT models for evaluation has become a common practice, their high cost and limited interpretability pose challenges for reproducible benchmarking.
These limitations have motivated our development of an LLM-free evaluation framework designed for open-ended, fine-grained video motion understanding.

We first develop a structured information extraction tool based on the NLTK library to obtain motion-related elements from the model's response, including camera motion, subject list, the motion lists of individual subjects, and the comprehensive temporal motion list (including actions of all subjects). 
This structured representation bridges model responses with quantitative evaluation metrics while maintaining human interpretability.

With the extracted elements, we calculate the score of the model's response through hierarchical sequence comparison.
Specifically, we adopt pre-trained models such as Sentence-BERT~\cite{reimers2019sentence} to calculate the semantic similarity of the extracted and manually annotated subject attributes and complete action sequences, and combine these two types of similarities for subject matching. 
Next, we conduct sequence comparisons in three aspects to obtain the comprehensive score: the camera motion sequence, the action sequence of each subject, and the comprehensive temporal action sequence. 
Taking the action sequence of one subject as an example, the predicted subject \(S\) is matched to the manually annotated subject \(G\).
Their action sequences are represented as $[a^s_1,...,a^s_n]$ and $[a^g_1,...,a^g_m]$ respectively. 
We construct an action similarity matrix \(\mathbf{M} \in \mathbb{R}^{n\times m}\) where each element $M_{ij}$ represents the semantic similarity between predicted action $a^s_i$ and ground truth action $a^g_j$:
\begingroup
\small
\begin{equation}
M_{ij} = \text{sim}(a^s_i, a^g_j)
\end{equation}
\endgroup
Based on this matrix and optimal matching, we calculate the similarity-weighted precision and recall:
\begingroup
\small
\begin{align}
P &= \frac{|\text{matched predicted actions}|}{|\text{predicted actions}|} \cdot \overline{\text{sim}} \cdot L_f, \\
R &= \frac{|\text{matched annotated actions}|}{|\text{annotated actions}|} \cdot \overline{\text{sim}} \cdot L_f,
\end{align}
\endgroup
where $\overline{\text{sim}}$ is the average similarity score of matched pairs.
$L_f$ is a length factor, which is used to penalize the unfair comparisons that may occur due to a large discrepancy in sequence lengths (such as numerous repeated descriptions). 
For each paired subjects, we further evaluate the order correctness using Kendall's Tau coefficient $\tau$ to measure the rank correlation between matched action indices. 
The calculated score for each subject pair is a weighted combination of multiple dimensions: $Score = w_p P + w_r R + w_o O$, where $w_s$, $w_r$ and $w_o$ are the weights of different indicators.
The camera motion and the comprehensive temporal action sequence can be scored similarly.
The final score of each model is obtained by averaging over all samples.

\renewcommand{\arraystretch}{1.2}
\begin{table*}[t]
\small
\resizebox{\linewidth}{!}{
\begin{tabular}{lcccccccccccc}
\toprule
%\specialrule{0em}{0.3pt}{0.3pt}
\multirow{2}{*}{\textbf{Methods}} & \multirow{2}{*}{\textbf{Date}} & \multirow{2}{*}{\textbf{Input}}  & \multicolumn{7}{c}{\textbf{Close-Ended}} & \multicolumn{3}{c}{\textbf{Open-Ended}}  \\ 
\specialrule{0em}{0.3pt}{0.3pt}
\cmidrule(r){4-10} \cmidrule(r){11-13} 
\specialrule{0em}{0.1pt}{0.1pt}
&~&~& ALL & AS & HAC & SAD & MAD & CM
& NSM  & GPT-C & GPT-D & LLM-Free \\
\specialrule{0em}{0.3pt}{0.3pt}
\hline
\rowcolor[HTML]{eff0f1}Full mark &-- &-- &100 &100  &100  &100  &100  &100  &100  &10 & 10  &100    \\
\rowcolor[HTML]{eff0f1}Random &-- &-- &20 &20 &20  &20  &20  &20  &20  &-- &-- &--    \\

\hline
 \rowcolor[HTML]{E3F8F8}\multicolumn{13}{l}{\textcolor{gray}{{\textit{\textbf{Proprietary MLLMs}}}}}\\
\rowcolor[HTML]{E3F8F8}Gemini-1.5-Pro~\cite{team2024gemini} & 2024-04 & 1 fps$^*$ & \bf{49.87} & \underline{49.22} & \bf{53.73} & \bf{48.80} & \bf{54.85} & \bf{41.58} & \underline{56.25} & \underline{4.52} & \bf{4.68} & \underline{52.91} \\
\rowcolor[HTML]{E3F8F8}GPT-4o~\cite{hurst2024gpt4o} & 2024-08 & 1 fps$^*$ & 42.09 & 40.65 & 45.10 & 42.84 & 45.48 & 36.00 & 48.44 & 4.33 & 4.01 & 49.50 \\
\rowcolor[HTML]{E3F8F8}Claude-3.7-Sonnet~\cite{Claude3.7} & 2025-02 & 1 fps$^*$ & 43.73 & 45.20 & 43.02 & 41.82 & 48.05 & 39.07 & 46.88 & 4.32 & \underline{4.63} & 43.03 \\
          
\hline
        
 \rowcolor[HTML]{FFF5F5}\multicolumn{13}{l}{\textcolor{gray}{{\textit{\textbf{Open-source MLLMs}}}}} \\

\rowcolor[HTML]{FFF5F5}Video-LLaVA-7B~\cite{lin2024video-llava} & 2023-11 & 8 frms & 25.37 & 24.91 & 21.54 & 25.45 & 30.54 & 26.23 & 21.88 & 2.18 & 2.31 & 41.36 \\
\rowcolor[HTML]{FFF5F5}LLaVA-NeXT-Video-7B~\cite{zhang2024llavanextvideo} & 2024-05 & 8 frms & 23.45 & 21.27 & 22.45 & 26.05 & 26.72 & 23.07 & 14.06 & 2.57 & 2.02 & 29.48 \\
\rowcolor[HTML]{FFF5F5}LLaVA-NeXT-Video-34B~\cite{zhang2024llavanextvideo} & 2024-05 & 8 frms & 30.44 & 31.70 & 31.99 & 32.31 & 22.99 & 29.58 & 46.88 & 2.83 & 2.67 & 39.41 \\
\rowcolor[HTML]{FFF5F5}Tarsier-7B~\cite{wang2024tarsier} & 2024-07 & 8 frms & 17.46 & 12.55 & 21.16 & 17.87 & 17.93 & 22.23 & 31.25 & 3.47 & 2.80 & 46.25 \\
\rowcolor[HTML]{FFF5F5}Tarsier-34B~\cite{wang2024tarsier} & 2024-07 & 8 frms & 30.34 & 28.56 & 34.98 & 26.90 & 31.29 & 31.91 & 37.50 & 3.79 & 2.97 & 47.13 \\
\rowcolor[HTML]{FFF5F5}Aria~\cite{aria} & 2024-10 & 8 frms & 34.63 & 33.33 & 41.14 & 30.14 & 35.27 & 33.21 & \bf{59.38} & 2.85 & 2.61 & 42.78 \\
\rowcolor[HTML]{FFF5F5}InternVL2.5-2B~\cite{chen2024internvl2.5} & 2024-12 & 8 frms & 22.90 & 18.70 & 28.23 & 23.71 & 27.47 & 19.16 & 23.44 & 2.80 & 2.99 & 43.23 \\
\rowcolor[HTML]{FFF5F5}InternVL2.5-8B~\cite{chen2024internvl2.5} & 2024-12 & 8 frms & 34.59 & 31.97 & 38.68 & 38.09 & 37.76 & 26.14 & 35.94 & 3.11 & 3.38 & 44.18 \\
\rowcolor[HTML]{FFF5F5}InternVL2.5-78B~\cite{chen2024internvl2.5} & 2024-12 & 8 frms & 38.54 & 38.38 & 40.62 & 39.05 & 43.65 & 29.40 & 39.06 & 2.98 & 3.41 & 44.01 \\
\rowcolor[HTML]{FFF5F5}Tarsier2-Recap-7B~\cite{yuan2025tarsier2} & 2024-12 & 16 frms & -- & -- & -- & -- & -- & -- & -- & \bf{4.60} & 4.38 & \bf{56.58} \\
\rowcolor[HTML]{FFF5F5}LLaVA-Video-7B-Qwen2~\cite{zhang2024llava-video} & 2024-10 & 64 frms & 38.60 & 36.14 & 41.27 & 41.28 & 44.48 & 29.58 & 46.88 & 3.57 & 3.40 & 45.41 \\
\rowcolor[HTML]{FFF5F5}LLaVA-Video-72B-Qwen2~\cite{zhang2024llava-video} & 2024-10 & 64 frms & 46.08 & 48.35 & 47.50 & 45.25 & 51.70 & 33.02 & 53.12 & 3.42 & 3.42 & 46.06 \\
\rowcolor[HTML]{FFF5F5}VideoChat-Flash-Qwen2-7B~\cite{li2024videochatflash} & 2025-01 & 1 fps & 43.82 & 41.90 & \underline{48.41} & 42.84 & 50.95 & 35.07 & 50.00 & 3.25 & 2.55 & 40.82 \\
\rowcolor[HTML]{FFF5F5}VideoLLaMA3-2B~\cite{zhang2025videollama3} & 2025-01 & 1 fps & 32.98 & 28.97 & 36.60 & 34.90 & 38.01 & 28.56 & 40.62 & 3.14 & 2.98 & 39.29 \\
\rowcolor[HTML]{FFF5F5}VideoLLaMA3-7B~\cite{zhang2025videollama3} & 2025-01 & 1 fps & 41.46 & 40.20 & 44.13 & 42.42 & 48.30 & 31.53 & 42.19 & 3.64 & 3.24 & 48.63 \\
\rowcolor[HTML]{FFF5F5}Qwen2.5-VL-3B~\cite{yang2024qwen2.5} & 2025-01 & 1 fps & 37.05 & 38.45 & 38.22 & 36.64 & 39.75 & 29.77 & 32.81 & 2.77 & 2.91 & 47.32 \\
\rowcolor[HTML]{FFF5F5}Qwen2.5-VL-7B~\cite{yang2024qwen2.5} & 2025-01 & 1 fps & 40.76 & 39.48 & 43.28 & 43.14 & 43.65 & 33.49 & 39.06 & 3.28 & 3.41 & 48.46 \\
\rowcolor[HTML]{FFF5F5}Qwen2.5-VL-72B~\cite{yang2024qwen2.5} & 2025-01 & 1 fps & \underline{48.14} & \bf{50.28} & 46.98 & \underline{48.13} & \underline{51.78} & \underline{40.28} & 51.56 & 3.37 & 3.44 & 49.72 \\

\hline

\rowcolor[HTML]{F1F6EC}Qwen2.5-VL-7B+FAVOR-Train & -- & 1 fps & 42.13 & 41.75 & 45.17 & 40.91 & 43.57 & 39.16 & 39.06 & 3.55 & 3.53 &  56.33 \\

\bottomrule
\end{tabular}
}
\vspace{-0.1cm}
\caption{The overall performances of 21 MLLMs on FAVOR-Bench, including close-ended multiple choice and open-ended evaluation with GPT-assisted and LLM-free scores. 
GPT-C and GPT-D mean correctness and detailedness scores generated by GPT-4o.
The highest and second-highest results among all MLLMs are indicated in bold and underlined.
Due to the API response limitations, the video input of proprietary MLLMs is restricted to 16 frames if the video is longer than 16 seconds (demoted as ``1 fps$^*$".) 
Tarsier2-Recap-7B is a model specially designed for captioning and it fails to fulfill the close-ended evaluation.} 
% \vspace{-3mm}
\label{tab:ourbench}
\end{table*}

\section{Experiments}
\label{sec:experiments}

\subsection{Experimental Settings}
We have conducted a comprehensive evaluation of 21 MLLMs through our FAVOR-Bench, which includes both open-source and proprietary models. For models that are part of a series, we evaluated their most recently released versions like VideoLLaMA3~\cite{zhang2025videollama3}, InternVL2.5~\cite{chen2024internvl2.5} and Qwen2.5-VL~\cite{yang2024qwen2.5}. The evaluation of all models is carried out using either their official implementations or accessible APIs, and all assessments are conducted in a zero-shot manner. We employ either a uniform sampling strategy or a frame rate sampling strategy to form the vision input following the official examples of each model. 
For close-ended evaluation, we prompt models to choose from the provided options rather than merely output the chosen indices. For open-ended evaluation, we prompt models to focus more on temporal dynamics rather than static contents.

\definecolor{Gray}{gray}{0.9}
\renewcommand{\arraystretch}{1.0}
\setlength{\tabcolsep}{4pt}
\begin{table*}[t]
\small
\resizebox{\linewidth}{!}{
\begin{tabular}{lccccccccccc|ccccccc}
\toprule
%\specialrule{0em}{0.3pt}{0.3pt}
\multirow{2}{*}{\textbf{Methods}} & \multicolumn{11}{c}{\textbf{TVBench}} & \multicolumn{7}{c}{\textbf{MotionBench-Dev}}  \\ 
\specialrule{0em}{0.3pt}{0.3pt}
\cmidrule(r){2-12} \cmidrule(r){13-19} 
\specialrule{0em}{0.1pt}{0.1pt}
& \cellcolor{Gray} AVG & AC & OC & AS & OS & ST & AL & AA & UA & ES & MD & \cellcolor{Gray} ALL & MR & LM & CM & MO & AO & RC\\     
\specialrule{0em}{0.3pt}{0.3pt}
\midrule

Random & \cellcolor{Gray} 33.3 & 25.0 & 25.0 & 50.0 & 33.3 & 50.0 & 25.0 & 50.0 & 25.0 & 25.0 & 25.0 & \cellcolor{Gray} 25.0 & 25.0 & 25.0 & 25.0 & 25.0 & 25.0 & 25.0  \\
 \midrule
Qwen2.5-VL-7B~\cite{yang2024qwen2.5} & \cellcolor{Gray} 45.2 & 36.8 & 36.5 & 64.8 & 37.3 & 62.7 & \bf{38.8} & 74.1 & \bf{41.5} & 26.0 & 33.6 & \cellcolor{Gray} 46.2 & 45.2 & \bf{45.4} & 42.3 & \bf{66.7} & 36.0 & \bf{33.0}\\

+ FAVOR-Train& \cellcolor{Gray} \bf{46.1} & \bf{43.1} & \bf{37.2} & \bf{68.0} & \bf{37.8} & \bf{62.7} & 34.4 & \bf{77.5} & 37.8 & \bf{29.0} & \bf{34.1} & \cellcolor{Gray} \bf{47.9} & \bf{49.2} & \bf{45.4} & \bf{43.1} & 66.2 & \bf{37.8} & 32.3\\

\bottomrule
 \end{tabular}
}
%\vspace{-0.2cm}
\caption{Comparison on TVBench and MotionBench with our proposed FAVOR-Train. AVG means the average score of all the 10 tasks of TVBench. ALL denotes the accuracy on all 4,018 questions of MotionBench-Dev. Qwen2.5-VL gains considerable performance improvement from fine-tuning with FAVOR-Train.} 
%\vspace{-0.5cm}
\label{tab:otherbench}
\end{table*}
\setlength{\tabcolsep}{6pt}

\subsection{Results Analysis on FAVOR-Bench}
We report the comprehensive results of the evaluated MLLMs on FAVOR-Bench in~\cref{tab:ourbench}, including the close-ended performance (per-task and overall scores) and the open-ended performance.
The results reveal critical limitations in fine-grained video motion understanding for existing MLLMs.

\noindent \textbf{Close-Ended Performance Analysis.}
As can be concluded from~\cref{tab:ourbench}, proprietary MLLMs generally demonstrate superior performance compared to open-source MLLMs, while this gap is narrowed by advanced open-source models.
Among all models, Gemini-1.5 achieves the highest overall score (49.87\%). 
However, this leading performance remains substantially below the practical deployment expectations for real-world applications.
For open-source models, we observe evident scaling effects.
For example, Qwen2.5-VL-72B achieves 48.14\%, outperforming its smaller variants by 8-11\%.
At the widely adopted 7B scale, VideoChat-Flash-Qwen2-7B~\cite{li2024videochatflash} achieved the highest performance (43.82\%). 
Notably, Tarsier2-Recap-7B specializes in caption generation tasks, so its close-ended performance is not compared.

We conduct task-specific analysis to bring further insights. 
First, most models achieved higher scores in the Holistic Action Classification task than their overall performance, confirming their competence in core action recognition.
Second, the Camera Motion task proves challenging, especially for models with strong comprehensive capabilities (such as Gemini-1.5, LLaVA-Video-72B-Qwen2 and Qwen2.5-VL-72B). Their CM scores are significantly lower than the overall scores, suggesting that MLLMs face difficulties in understanding viewpoint dynamics and focus shifts.
Third, MLLMs exhibit stronger Multi-Action Details understanding capabilities over Single-Action Detail capabilities, indicating that MLLMs are better at aggregating and analyzing temporal information compared to capturing details at specific instances.

\noindent \textbf{Open-Ended Performance Analysis.}
The open-ended evaluation results in Table~\ref{tab:ourbench} expose critical limitations in current MLLMs' ability to understand and descibe fine-grained motions in videos.
The evaluated models did not score highly in both types of evaluations, among which Claude-3.7-Sonnet and Tarsier2-Recap-7B achieve the highest performance in the GPT-assisted evaluation and the LLM-free evaluation respectively. Claude focuses more on contents of each frame in its output captions, rather than conveying coherent action information that spans across frames. Consequently, this results in lower alignment between the described actions and the ground truth (GT) actions. Tarsier2-Recap-7B fails to fulfill the close-ended evaluation, and it is not suitable for other tasks such as text generation, dialogue, and reasoning. However, it performs very well on video caption tasks, even outperforming proprietary models.
In addition, the scaling effects can also be observed. 
For example, the overall performance of VideoLLaMA3-7B (3.64 / 3.24 / 48.63) is superior to that of VideoLLaMA3-3B (3.14 / 2.98 / 39.29). This indicates that as the model scale increases, its fine-grained motion description capabilities also improve. 
More detailed results of the LLM-free evaluation are provided in the supplementary material and shown in~\ref{tab:llm-free-detail}.

\subsection{FAVOR-Train Set}
To facilitate better video motion understanding and description, we further propose a training set FAVOR-Train including 25K videos and the corresponding manual captions. This dataset is divided into two parts: ego-centric (first-person) and third-person views. All the 14,038 third-person view videos are sourced from the Koala36M~\cite{wang2024koala_dataset} dataset, which provides a rich variety of motions and interactions in vast scenarios. For the ego-centric portion with 3,114 videos, we curated data from four distinct datasets EgoTaskQA~\cite{jia2022egotaskqa}, Charades-Ego~\cite{sigurdsson2018charadesego}, EgoExo4D~\cite{grauman2024ego} and EgoExoLearn~\cite{huang2024egoexolearn}. To ensure a diverse range of activities and contexts within our dataset, specially designed sampling strategies are employed. There is no intersection between the videos in FAVOR-Train and FAVOR-Bench. Details of the sampling strategies can be found in the supplementary material.

To validate the effectiveness of FAVOR-Train, we fine-tune the Qwen2.5-VL model using FAVOR-Train data. 
We then report the performance on both the FAVOR-Bench and existing benchmarks with motion-related tasks.
Specifically, we conduct the validation on TVBench~\cite{cores2024tvbench} and MotionBench~\cite{hong2025motionbench}. 
As shown in the last lines in ~\cref{tab:ourbench}, FAVOR-Train brings a $1.37\%$ accuracy gain in our close-ended tasks. 
In our open-ended evaluation, the proposed training set can significantly boost video caption performance. It improves the LLM-Free score by $7.9\%$ and GPT evaluation score by $0.2$. After training with FAVOR-Train set, the model can output more fine-grained descriptions of action and camera motion, like bending down, turning head and camera shaking. Existence of these descriptions brings it a higher matching score in LLM-free evaluation. Besides, lacking of descriptions of static details, appearance and environments and the short in lengths restricted further improvement of GPT-assisted scores.
\Cref{tab:otherbench} demonstrates that our FAVOR-Train can also effectively improve the overall performance of the baseline Qwen2.5-VL-7B on existing benchmarks, especially for motion-related tasks including Action Count (AC), Action Sequence (AS), Egocentric Sequence (ES) and Motion Recognition(MR).

% \Cref{tab:otherbench} shows that our FAVOR - Train can help improve the baseline performance in previous benchmarks, especially in action - related tasks such as Action Count (AC), Action Sequence (AS), Egocentric Sequence (ES), and Motion Recognition (MR).

% Our FAVOR-Train fine-tuned model shows meaningful improvements over its base version, with overall accuracy increasing from 40.76\% to 42.13\% and notable gains in the difficult Camera Motion task (33.49\% to 39.16\%).

% 数据规模
% 训练过程
% 效果（FAVOR）
% 其他benchmark效果（不掉点）

\section{Conclusion}
\label{sec:conclusion}

We present FAVOR-Bench, the first comprehensive benchmark for evaluating fine-grained video motion understanding in MLLMs, featuring 1,776 diverse videos with both close-ended (8,184 multiple-choice QA pairs across six distinct tasks) and open-ended (GPT-assist and a novel LLM-free framework) evaluation.
The comprehensive results of 21 state-of-the-art MLLMs reveal significant limitations in understanding detailed temporal dynamics.
To narrow this gap, we further construct FAVOR-Train with 17,279 annotated videos, which effectively improves the motion understanding performance both on our proposed FAVOR-Bench and motion-related tasks of existing benchmarks.
Through FAVOR-Bench and FAVOR-Train, we provide valuable tools to the community for the development of more powerful video understanding models.

{
    \small
    \bibliographystyle{ieeenat_fullname}
    \bibliography{main}

\begin{thebibliography}{55}
\providecommand{\natexlab}[1]{#1}
\providecommand{\url}[1]{\texttt{#1}}
\expandafter\ifx\csname urlstyle\endcsname\relax
  \providecommand{\doi}[1]{doi: #1}\else
  \providecommand{\doi}{doi: \begingroup \urlstyle{rm}\Url}\fi

\bibitem[Anthropic(2025)]{Claude3.7}
Anthropic.
\newblock Claude 3.7.
\newblock \url{https://www.anthropic.com/claude/sonnet}, 2025.

\bibitem[Athar et~al.(2024)Athar, Deng, and Chen]{athar2024vicas_task}
Ali Athar, Xueqing Deng, and Liang-Chieh Chen.
\newblock Vicas: A dataset for combining holistic and pixel-level video understanding using captions with grounded segmentation.
\newblock \emph{arXiv preprint arXiv:2412.09754}, 2024.

\bibitem[Caba~Heilbron et~al.(2015)Caba~Heilbron, Escorcia, Ghanem, and Carlos~Niebles]{caba2015activitynet}
Fabian Caba~Heilbron, Victor Escorcia, Bernard Ghanem, and Juan Carlos~Niebles.
\newblock Activitynet: A large-scale video benchmark for human activity understanding.
\newblock In \emph{Proceedings of the ieee conference on computer vision and pattern recognition}, pages 961--970, 2015.

\bibitem[Chen et~al.(2024{\natexlab{a}})Chen, Siarohin, Menapace, Deyneka, Chao, Jeon, Fang, Lee, Ren, Yang, et~al.]{chen2024panda_dataset}
Tsai-Shien Chen, Aliaksandr Siarohin, Willi Menapace, Ekaterina Deyneka, Hsiang-wei Chao, Byung~Eun Jeon, Yuwei Fang, Hsin-Ying Lee, Jian Ren, Ming-Hsuan Yang, et~al.
\newblock Panda-70m: Captioning 70m videos with multiple cross-modality teachers.
\newblock In \emph{Proceedings of the IEEE/CVF Conference on Computer Vision and Pattern Recognition}, pages 13320--13331, 2024{\natexlab{a}}.

\bibitem[Chen et~al.(2024{\natexlab{b}})Chen, Wang, Cao, Liu, Gao, Cui, Zhu, Ye, Tian, Liu, et~al.]{chen2024internvl2.5}
Zhe Chen, Weiyun Wang, Yue Cao, Yangzhou Liu, Zhangwei Gao, Erfei Cui, Jinguo Zhu, Shenglong Ye, Hao Tian, Zhaoyang Liu, et~al.
\newblock Expanding performance boundaries of open-source multimodal models with model, data, and test-time scaling.
\newblock \emph{arXiv preprint arXiv:2412.05271}, 2024{\natexlab{b}}.

\bibitem[Chen et~al.(2024{\natexlab{c}})Chen, Wang, Tian, Ye, Gao, Cui, Tong, Hu, Luo, Ma, et~al.]{chen2024internvl1.5}
Zhe Chen, Weiyun Wang, Hao Tian, Shenglong Ye, Zhangwei Gao, Erfei Cui, Wenwen Tong, Kongzhi Hu, Jiapeng Luo, Zheng Ma, et~al.
\newblock How far are we to gpt-4v? closing the gap to commercial multimodal models with open-source suites.
\newblock \emph{Science China Information Sciences}, 67\penalty0 (12):\penalty0 220101, 2024{\natexlab{c}}.

\bibitem[Cheng et~al.(2024)Cheng, Leng, Zhang, Xin, Li, Chen, Zhu, Zhang, Luo, Zhao, et~al.]{cheng2024videollama2}
Zesen Cheng, Sicong Leng, Hang Zhang, Yifei Xin, Xin Li, Guanzheng Chen, Yongxin Zhu, Wenqi Zhang, Ziyang Luo, Deli Zhao, et~al.
\newblock Videollama 2: Advancing spatial-temporal modeling and audio understanding in video-llms.
\newblock \emph{arXiv preprint arXiv:2406.07476}, 2024.

\bibitem[Cores et~al.(2024)Cores, Dorkenwald, Mucientes, Snoek, and Asano]{cores2024tvbench}
Daniel Cores, Michael Dorkenwald, Manuel Mucientes, Cees~GM Snoek, and Yuki~M Asano.
\newblock Tvbench: Redesigning video-language evaluation.
\newblock \emph{arXiv preprint arXiv:2410.07752}, 2024.

\bibitem[Cui et~al.(2023)Cui, Wang, Shafiullah, and Pinto]{cui2023play_mimic}
Zichen~Jeff Cui, Yibin Wang, Nur~Muhammad Shafiullah, and Lerrel Pinto.
\newblock From play to policy: Conditional behavior generation from uncurated robot data.
\newblock In \emph{11th International Conference on Learning Representations, ICLR 2023}, 2023.

\bibitem[Fu et~al.(2024)Fu, Dai, Luo, Li, Ren, Zhang, Wang, Zhou, Shen, Zhang, et~al.]{fu2024video-mme}
Chaoyou Fu, Yuhan Dai, Yongdong Luo, Lei Li, Shuhuai Ren, Renrui Zhang, Zihan Wang, Chenyu Zhou, Yunhang Shen, Mengdan Zhang, et~al.
\newblock Video-mme: The first-ever comprehensive evaluation benchmark of multi-modal llms in video analysis.
\newblock \emph{arXiv preprint arXiv:2405.21075}, 2024.

\bibitem[Grauman et~al.(2024)Grauman, Westbury, Torresani, Kitani, Malik, Afouras, Ashutosh, Baiyya, Bansal, Boote, et~al.]{grauman2024ego}
Kristen Grauman, Andrew Westbury, Lorenzo Torresani, Kris Kitani, Jitendra Malik, Triantafyllos Afouras, Kumar Ashutosh, Vijay Baiyya, Siddhant Bansal, Bikram Boote, et~al.
\newblock Ego-exo4d: Understanding skilled human activity from first-and third-person perspectives.
\newblock In \emph{Proceedings of the IEEE/CVF Conference on Computer Vision and Pattern Recognition}, pages 19383--19400, 2024.

\bibitem[Guo et~al.(2025)Guo, Yang, Zhang, Song, Zhang, Xu, Zhu, Ma, Wang, Bi, et~al.]{guo2025deepseek}
Daya Guo, Dejian Yang, Haowei Zhang, Junxiao Song, Ruoyu Zhang, Runxin Xu, Qihao Zhu, Shirong Ma, Peiyi Wang, Xiao Bi, et~al.
\newblock Deepseek-r1: Incentivizing reasoning capability in llms via reinforcement learning.
\newblock \emph{arXiv preprint arXiv:2501.12948}, 2025.

\bibitem[Hong et~al.(2025)Hong, Cheng, Yang, Wang, Wang, Gu, Huang, Dong, and Tang]{hong2025motionbench}
Wenyi Hong, Yean Cheng, Zhuoyi Yang, Weihan Wang, Lefan Wang, Xiaotao Gu, Shiyu Huang, Yuxiao Dong, and Jie Tang.
\newblock Motionbench: Benchmarking and improving fine-grained video motion understanding for vision language models.
\newblock \emph{arXiv preprint arXiv:2501.02955}, 2025.

\bibitem[Hu et~al.(2022)Hu, Luo, and Chen]{hu2022make_ti2v}
Yaosi Hu, Chong Luo, and Zhenzhong Chen.
\newblock Make it move: controllable image-to-video generation with text descriptions.
\newblock In \emph{Proceedings of the IEEE/CVF Conference on Computer Vision and Pattern Recognition}, pages 18219--18228, 2022.

\bibitem[Huang et~al.(2018)Huang, Ramanathan, Mahajan, Torresani, Paluri, Fei-Fei, and Niebles]{huang2018makes_single_frame}
De-An Huang, Vignesh Ramanathan, Dhruv Mahajan, Lorenzo Torresani, Manohar Paluri, Li Fei-Fei, and Juan~Carlos Niebles.
\newblock What makes a video a video: Analyzing temporal information in video understanding models and datasets.
\newblock In \emph{Proceedings of the IEEE Conference on Computer Vision and Pattern Recognition}, pages 7366--7375, 2018.

\bibitem[Huang et~al.(2024{\natexlab{a}})Huang, Chen, Xu, Zhang, Yang, Pei, Zhang, Dong, Wang, Wang, et~al.]{huang2024egoexolearn}
Yifei Huang, Guo Chen, Jilan Xu, Mingfang Zhang, Lijin Yang, Baoqi Pei, Hongjie Zhang, Lu Dong, Yali Wang, Limin Wang, et~al.
\newblock Egoexolearn: A dataset for bridging asynchronous ego-and exo-centric view of procedural activities in real world.
\newblock In \emph{Proceedings of the IEEE/CVF Conference on Computer Vision and Pattern Recognition}, pages 22072--22086, 2024{\natexlab{a}}.

\bibitem[Huang et~al.(2024{\natexlab{b}})Huang, He, Yu, Zhang, Si, Jiang, Zhang, Wu, Jin, Chanpaisit, et~al.]{huang2024vbench_task}
Ziqi Huang, Yinan He, Jiashuo Yu, Fan Zhang, Chenyang Si, Yuming Jiang, Yuanhan Zhang, Tianxing Wu, Qingyang Jin, Nattapol Chanpaisit, et~al.
\newblock Vbench: Comprehensive benchmark suite for video generative models.
\newblock In \emph{Proceedings of the IEEE/CVF Conference on Computer Vision and Pattern Recognition}, pages 21807--21818, 2024{\natexlab{b}}.

\bibitem[Hurst et~al.(2024)Hurst, Lerer, Goucher, Perelman, Ramesh, Clark, Ostrow, Welihinda, Hayes, Radford, et~al.]{hurst2024gpt4o}
Aaron Hurst, Adam Lerer, Adam~P Goucher, Adam Perelman, Aditya Ramesh, Aidan Clark, AJ Ostrow, Akila Welihinda, Alan Hayes, Alec Radford, et~al.
\newblock Gpt-4o system card.
\newblock \emph{arXiv preprint arXiv:2410.21276}, 2024.

\bibitem[Jia et~al.(2022)Jia, Lei, Zhu, and Huang]{jia2022egotaskqa}
Baoxiong Jia, Ting Lei, Song-Chun Zhu, and Siyuan Huang.
\newblock Egotaskqa: Understanding human tasks in egocentric videos.
\newblock \emph{Advances in Neural Information Processing Systems}, 35:\penalty0 3343--3360, 2022.

\bibitem[Kay et~al.(2017)Kay, Carreira, Simonyan, Zhang, Hillier, Vijayanarasimhan, Viola, Green, Back, Natsev, et~al.]{kay2017kinetics}
Will Kay, Joao Carreira, Karen Simonyan, Brian Zhang, Chloe Hillier, Sudheendra Vijayanarasimhan, Fabio Viola, Tim Green, Trevor Back, Paul Natsev, et~al.
\newblock The kinetics human action video dataset.
\newblock \emph{arXiv preprint arXiv:1705.06950}, 2017.

\bibitem[Lei et~al.(2023)Lei, Berg, and Bansal]{lei2023revealing_single_frame}
Jie Lei, Tamara Berg, and Mohit Bansal.
\newblock Revealing single frame bias for video-and-language learning.
\newblock In \emph{Proceedings of the 61st Annual Meeting of the Association for Computational Linguistics (Volume 1: Long Papers)}, pages 487--507, 2023.

\bibitem[Li et~al.(2024{\natexlab{a}})Li, Liu, Wu, Wang, Shen, Qu, Niu, Wang, Chen, and Li]{aria}
Dongxu Li, Yudong Liu, Haoning Wu, Yue Wang, Zhiqi Shen, Bowen Qu, Xinyao Niu, Guoyin Wang, Bei Chen, and Junnan Li.
\newblock Aria: An open multimodal native mixture-of-experts model.
\newblock \emph{arXiv preprint arXiv:2410.05993}, 2024{\natexlab{a}}.

\bibitem[Li et~al.(2024{\natexlab{b}})Li, Wang, He, Li, Wang, Liu, Wang, Xu, Chen, Luo, et~al.]{li2024mvbench}
Kunchang Li, Yali Wang, Yinan He, Yizhuo Li, Yi Wang, Yi Liu, Zun Wang, Jilan Xu, Guo Chen, Ping Luo, et~al.
\newblock Mvbench: A comprehensive multi-modal video understanding benchmark.
\newblock In \emph{Proceedings of the IEEE/CVF Conference on Computer Vision and Pattern Recognition}, pages 22195--22206, 2024{\natexlab{b}}.

\bibitem[Li et~al.(2024{\natexlab{c}})Li, Wang, Yu, Zeng, Zhu, Huang, Gao, Li, He, Wang, et~al.]{li2024videochatflash}
Xinhao Li, Yi Wang, Jiashuo Yu, Xiangyu Zeng, Yuhan Zhu, Haian Huang, Jianfei Gao, Kunchang Li, Yinan He, Chenting Wang, et~al.
\newblock Videochat-flash: Hierarchical compression for long-context video modeling.
\newblock \emph{arXiv preprint arXiv:2501.00574}, 2024{\natexlab{c}}.

\bibitem[Li et~al.(2024{\natexlab{d}})Li, Chen, Hu, Wang, Shi, and Zhang]{li2024videovista_task}
Yunxin Li, Xinyu Chen, Baotian Hu, Longyue Wang, Haoyuan Shi, and Min Zhang.
\newblock Videovista: A versatile benchmark for video understanding and reasoning.
\newblock \emph{arXiv preprint arXiv:2406.11303}, 2024{\natexlab{d}}.

\bibitem[Lin et~al.(2024)Lin, Ye, Zhu, Cui, Ning, Jin, and Yuan]{lin2024video-llava}
Bin Lin, Yang Ye, Bin Zhu, Jiaxi Cui, Munan Ning, Peng Jin, and Li Yuan.
\newblock Video-llava: Learning united visual representation by alignment before projection.
\newblock In \emph{Proceedings of the 2024 Conference on Empirical Methods in Natural Language Processing}, pages 5971--5984, 2024.

\bibitem[Mangalam et~al.(2023)Mangalam, Akshulakov, and Malik]{mangalam2023egoschema}
Karttikeya Mangalam, Raiymbek Akshulakov, and Jitendra Malik.
\newblock Egoschema: A diagnostic benchmark for very long-form video language understanding.
\newblock \emph{Advances in Neural Information Processing Systems}, 36:\penalty0 46212--46244, 2023.

\bibitem[Monfort et~al.(2019)Monfort, Andonian, Zhou, Ramakrishnan, Bargal, Yan, Brown, Fan, Gutfreund, Vondrick, et~al.]{monfort2019moments}
Mathew Monfort, Alex Andonian, Bolei Zhou, Kandan Ramakrishnan, Sarah~Adel Bargal, Tom Yan, Lisa Brown, Quanfu Fan, Dan Gutfreund, Carl Vondrick, et~al.
\newblock Moments in time dataset: one million videos for event understanding.
\newblock \emph{IEEE transactions on pattern analysis and machine intelligence}, 42\penalty0 (2):\penalty0 502--508, 2019.

\bibitem[Nan et~al.(2024)Nan, Xie, Zhou, Fan, Yang, Chen, Li, Yang, and Tai]{nan2024openvid_dataset}
Kepan Nan, Rui Xie, Penghao Zhou, Tiehan Fan, Zhenheng Yang, Zhijie Chen, Xiang Li, Jian Yang, and Ying Tai.
\newblock Openvid-1m: A large-scale high-quality dataset for text-to-video generation.
\newblock \emph{arXiv preprint arXiv:2407.02371}, 2024.

\bibitem[Ni et~al.(2024)Ni, Egger, Lohit, Cherian, Wang, Koike-Akino, Huang, and Marks]{ni2024ti2v}
Haomiao Ni, Bernhard Egger, Suhas Lohit, Anoop Cherian, Ye Wang, Toshiaki Koike-Akino, Sharon~X Huang, and Tim~K Marks.
\newblock Ti2v-zero: Zero-shot image conditioning for text-to-video diffusion models.
\newblock In \emph{Proceedings of the IEEE/CVF Conference on Computer Vision and Pattern Recognition}, pages 9015--9025, 2024.

\bibitem[Reimers and Gurevych(2019)]{reimers2019sentence}
Nils Reimers and Iryna Gurevych.
\newblock Sentence-bert: Sentence embeddings using siamese bert-networks.
\newblock In \emph{Proceedings of the 2019 Conference on Empirical Methods in Natural Language Processing and the 9th International Joint Conference on Natural Language Processing (EMNLP-IJCNLP)}, pages 3982--3992, 2019.

\bibitem[Sevilla-Lara et~al.(2021)Sevilla-Lara, Zha, Yan, Goswami, Feiszli, and Torresani]{sevilla2021only_time_can_tell}
Laura Sevilla-Lara, Shengxin Zha, Zhicheng Yan, Vedanuj Goswami, Matt Feiszli, and Lorenzo Torresani.
\newblock Only time can tell: Discovering temporal data for temporal modeling.
\newblock In \emph{Proceedings of the IEEE/CVF winter conference on applications of computer vision}, pages 535--544, 2021.

\bibitem[Sigurdsson et~al.(2016)Sigurdsson, Varol, Wang, Farhadi, Laptev, and Gupta]{sigurdsson2016hollywood_charades}
Gunnar~A Sigurdsson, G{\"u}l Varol, Xiaolong Wang, Ali Farhadi, Ivan Laptev, and Abhinav Gupta.
\newblock Hollywood in homes: Crowdsourcing data collection for activity understanding.
\newblock In \emph{Computer Vision--ECCV 2016: 14th European Conference, Amsterdam, The Netherlands, October 11--14, 2016, Proceedings, Part I 14}, pages 510--526. Springer, 2016.

\bibitem[Sigurdsson et~al.(2018)Sigurdsson, Gupta, Schmid, Farhadi, and Alahari]{sigurdsson2018charadesego}
Gunnar~A Sigurdsson, Abhinav Gupta, Cordelia Schmid, Ali Farhadi, and Karteek Alahari.
\newblock Charades-ego: A large-scale dataset of paired third and first person videos.
\newblock \emph{arXiv preprint arXiv:1804.09626}, 2018.

\bibitem[Song et~al.(2024)Song, Chai, Wang, Zhang, Zhou, Wu, Chi, Guo, Ye, Zhang, et~al.]{song2024moviechat}
Enxin Song, Wenhao Chai, Guanhong Wang, Yucheng Zhang, Haoyang Zhou, Feiyang Wu, Haozhe Chi, Xun Guo, Tian Ye, Yanting Zhang, et~al.
\newblock Moviechat: From dense token to sparse memory for long video understanding.
\newblock In \emph{Proceedings of the IEEE/CVF Conference on Computer Vision and Pattern Recognition}, pages 18221--18232, 2024.

\bibitem[Sun et~al.(2024)Sun, Shen, Cao, Liu, Li, Shen, Gan, Gui, Wang, Yang, et~al.]{sun2024aligning_gpt_assist}
Zhiqing Sun, Sheng Shen, Shengcao Cao, Haotian Liu, Chunyuan Li, Yikang Shen, Chuang Gan, Liang~Yan Gui, Yu~Xiong Wang, Yiming Yang, et~al.
\newblock Aligning large multimodal models with factually augmented rlhf.
\newblock In \emph{Findings of the 62nd Annual Meeting of the Association for Computational Linguistics, ACL 2024}, pages 13088--13110. Association for Computational Linguistics (ACL), 2024.

\bibitem[Team et~al.(2024)Team, Georgiev, Lei, Burnell, Bai, Gulati, Tanzer, Vincent, Pan, Wang, et~al.]{team2024gemini}
Gemini Team, Petko Georgiev, Ving~Ian Lei, Ryan Burnell, Libin Bai, Anmol Gulati, Garrett Tanzer, Damien Vincent, Zhufeng Pan, Shibo Wang, et~al.
\newblock Gemini 1.5: Unlocking multimodal understanding across millions of tokens of context.
\newblock \emph{arXiv preprint arXiv:2403.05530}, 2024.

\bibitem[Wang et~al.(2023)Wang, Fan, Sun, Zhang, Fei-Fei, Xu, Zhu, and Anandkumar]{wang2023mimicplay}
Chen Wang, Linxi Fan, Jiankai Sun, Ruohan Zhang, Li Fei-Fei, Danfei Xu, Yuke Zhu, and Anima Anandkumar.
\newblock Mimicplay: Long-horizon imitation learning by watching human play.
\newblock In \emph{Conference on Robot Learning}, pages 201--221. PMLR, 2023.

\bibitem[Wang et~al.(2024{\natexlab{a}})Wang, Yuan, Zhang, and Sun]{wang2024tarsier}
Jiawei Wang, Liping Yuan, Yuchen Zhang, and Haomiao Sun.
\newblock Tarsier: Recipes for training and evaluating large video description models.
\newblock \emph{arXiv preprint arXiv:2407.00634}, 2024{\natexlab{a}}.

\bibitem[Wang et~al.(2024{\natexlab{b}})Wang, Bai, Tan, Wang, Fan, Bai, Chen, Liu, Wang, Ge, et~al.]{wang2024qwen2}
Peng Wang, Shuai Bai, Sinan Tan, Shijie Wang, Zhihao Fan, Jinze Bai, Keqin Chen, Xuejing Liu, Jialin Wang, Wenbin Ge, et~al.
\newblock Qwen2-vl: Enhancing vision-language model's perception of the world at any resolution.
\newblock \emph{arXiv preprint arXiv:2409.12191}, 2024{\natexlab{b}}.

\bibitem[Wang et~al.(2024{\natexlab{c}})Wang, Shi, Ou, Chen, Lin, Wang, Jiang, Yang, Zheng, Tao, et~al.]{wang2024koala_dataset}
Qiuheng Wang, Yukai Shi, Jiarong Ou, Rui Chen, Ke Lin, Jiahao Wang, Boyuan Jiang, Haotian Yang, Mingwu Zheng, Xin Tao, et~al.
\newblock Koala-36m: A large-scale video dataset improving consistency between fine-grained conditions and video content.
\newblock \emph{arXiv preprint arXiv:2410.08260}, 2024{\natexlab{c}}.

\bibitem[Wu et~al.(2024)Wu, Yu, Chen, Tenenbaum, and Gan]{wu2024star}
Bo Wu, Shoubin Yu, Zhenfang Chen, Joshua~B Tenenbaum, and Chuang Gan.
\newblock Star: A benchmark for situated reasoning in real-world videos.
\newblock \emph{arXiv preprint arXiv:2405.09711}, 2024.

\bibitem[Xiao et~al.(2021)Xiao, Shang, Yao, and Chua]{xiao2021next-qa}
Junbin Xiao, Xindi Shang, Angela Yao, and Tat-Seng Chua.
\newblock Next-qa: Next phase of question-answering to explaining temporal actions.
\newblock In \emph{Proceedings of the IEEE/CVF conference on computer vision and pattern recognition}, pages 9777--9786, 2021.

\bibitem[Xie et~al.(2024)Xie, Zhang, Zhou, Li, Zhang, Hessel, Yang, and Liu]{xie2024funqa}
Binzhu Xie, Sicheng Zhang, Zitang Zhou, Bo Li, Yuanhan Zhang, Jack Hessel, Jingkang Yang, and Ziwei Liu.
\newblock Funqa: Towards surprising video comprehension.
\newblock In \emph{European Conference on Computer Vision}, pages 39--57. Springer, 2024.

\bibitem[Xu et~al.(2016)Xu, Mei, Yao, and Rui]{xu2016msr-vtt}
Jun Xu, Tao Mei, Ting Yao, and Yong Rui.
\newblock Msr-vtt: A large video description dataset for bridging video and language.
\newblock In \emph{Proceedings of the IEEE conference on computer vision and pattern recognition}, pages 5288--5296, 2016.

\bibitem[Yang et~al.(2024{\natexlab{a}})Yang, Yang, Hui, Zheng, Yu, Zhou, Li, Li, Liu, Huang, Dong, Wei, Lin, Tang, Wang, Yang, Tu, Zhang, Ma, Yang, Xu, Zhou, Bai, He, Lin, Dang, Lu, Chen, Yang, Li, Xue, Ni, Zhang, Wang, Peng, Men, Gao, Lin, Wang, Bai, Tan, Zhu, Li, Liu, Ge, Deng, Zhou, Ren, Zhang, Wei, Ren, Liu, Fan, Yao, Zhang, Wan, Chu, Liu, Cui, Zhang, Guo, and Fan]{yang2024qwen2technicalreport}
An Yang, Baosong Yang, Binyuan Hui, Bo Zheng, Bowen Yu, Chang Zhou, Chengpeng Li, Chengyuan Li, Dayiheng Liu, Fei Huang, Guanting Dong, Haoran Wei, Huan Lin, Jialong Tang, Jialin Wang, Jian Yang, Jianhong Tu, Jianwei Zhang, Jianxin Ma, Jianxin Yang, Jin Xu, Jingren Zhou, Jinze Bai, Jinzheng He, Junyang Lin, Kai Dang, Keming Lu, Keqin Chen, Kexin Yang, Mei Li, Mingfeng Xue, Na Ni, Pei Zhang, Peng Wang, Ru Peng, Rui Men, Ruize Gao, Runji Lin, Shijie Wang, Shuai Bai, Sinan Tan, Tianhang Zhu, Tianhao Li, Tianyu Liu, Wenbin Ge, Xiaodong Deng, Xiaohuan Zhou, Xingzhang Ren, Xinyu Zhang, Xipin Wei, Xuancheng Ren, Xuejing Liu, Yang Fan, Yang Yao, Yichang Zhang, Yu Wan, Yunfei Chu, Yuqiong Liu, Zeyu Cui, Zhenru Zhang, Zhifang Guo, and Zhihao Fan.
\newblock Qwen2 technical report, 2024{\natexlab{a}}.

\bibitem[Yang et~al.(2024{\natexlab{b}})Yang, Yang, Zhang, Hui, Zheng, Yu, Li, Liu, Huang, Wei, et~al.]{yang2024qwen2.5}
An Yang, Baosong Yang, Beichen Zhang, Binyuan Hui, Bo Zheng, Bowen Yu, Chengyuan Li, Dayiheng Liu, Fei Huang, Haoran Wei, et~al.
\newblock Qwen2. 5 technical report.
\newblock \emph{arXiv preprint arXiv:2412.15115}, 2024{\natexlab{b}}.

\bibitem[Yin et~al.(2023)Yin, Wang, Cao, Shi, Liu, Li, Huang, Wang, Sheng, Bai, et~al.]{yin2023lamm_task}
Zhenfei Yin, Jiong Wang, Jianjian Cao, Zhelun Shi, Dingning Liu, Mukai Li, Xiaoshui Huang, Zhiyong Wang, Lu Sheng, Lei Bai, et~al.
\newblock Lamm: Language-assisted multi-modal instruction-tuning dataset, framework, and benchmark.
\newblock \emph{Advances in Neural Information Processing Systems}, 36:\penalty0 26650--26685, 2023.

\bibitem[Yu et~al.(2019)Yu, Xu, Yu, Yu, Zhao, Zhuang, and Tao]{yu2019activitynet-qa}
Zhou Yu, Dejing Xu, Jun Yu, Ting Yu, Zhou Zhao, Yueting Zhuang, and Dacheng Tao.
\newblock Activitynet-qa: A dataset for understanding complex web videos via question answering.
\newblock In \emph{Proceedings of the AAAI Conference on Artificial Intelligence}, pages 9127--9134, 2019.

\bibitem[Yuan et~al.(2025)Yuan, Wang, Sun, Zhang, and Lin]{yuan2025tarsier2}
Liping Yuan, Jiawei Wang, Haomiao Sun, Yuchen Zhang, and Yuan Lin.
\newblock Tarsier2: Advancing large vision-language models from detailed video description to comprehensive video understanding.
\newblock \emph{arXiv preprint arXiv:2501.07888}, 2025.

\bibitem[Zhang et~al.(2025)Zhang, Li, Cheng, Hu, Yuan, Chen, Leng, Jiang, Zhang, Li, et~al.]{zhang2025videollama3}
Boqiang Zhang, Kehan Li, Zesen Cheng, Zhiqiang Hu, Yuqian Yuan, Guanzheng Chen, Sicong Leng, Yuming Jiang, Hang Zhang, Xin Li, et~al.
\newblock Videollama 3: Frontier multimodal foundation models for image and video understanding.
\newblock \emph{arXiv preprint arXiv:2501.13106}, 2025.

\bibitem[Zhang et~al.(2024{\natexlab{a}})Zhang, Li, Liu, Lee, Gui, Fu, Feng, Liu, and Li]{zhang2024llavanextvideo}
Yuanhan Zhang, Bo Li, haotian Liu, Yong~jae Lee, Liangke Gui, Di Fu, Jiashi Feng, Ziwei Liu, and Chunyuan Li.
\newblock Llava-next: A strong zero-shot video understanding model, 2024{\natexlab{a}}.

\bibitem[Zhang et~al.(2024{\natexlab{b}})Zhang, Wu, Li, Li, Ma, Liu, and Li]{zhang2024llava-video}
Yuanhan Zhang, Jinming Wu, Wei Li, Bo Li, Zejun Ma, Ziwei Liu, and Chunyuan Li.
\newblock Video instruction tuning with synthetic data, 2024{\natexlab{b}}.

\bibitem[Zhou et~al.(2024{\natexlab{a}})Zhou, Shu, Zhao, Wu, Xiao, Yang, Xiong, Zhang, Huang, and Liu]{zhou2024mlvu}
Junjie Zhou, Yan Shu, Bo Zhao, Boya Wu, Shitao Xiao, Xi Yang, Yongping Xiong, Bo Zhang, Tiejun Huang, and Zheng Liu.
\newblock Mlvu: A comprehensive benchmark for multi-task long video understanding.
\newblock \emph{arXiv preprint arXiv:2406.04264}, 2024{\natexlab{a}}.

\bibitem[Zhou et~al.(2024{\natexlab{b}})Zhou, Arnab, Buch, Yan, Myers, Xiong, Nagrani, and Schmid]{zhou2024streaming_task}
Xingyi Zhou, Anurag Arnab, Shyamal Buch, Shen Yan, Austin Myers, Xuehan Xiong, Arsha Nagrani, and Cordelia Schmid.
\newblock Streaming dense video captioning.
\newblock In \emph{Proceedings of the IEEE/CVF Conference on Computer Vision and Pattern Recognition}, pages 18243--18252, 2024{\natexlab{b}}.

\end{thebibliography}
}

\clearpage
\setcounter{page}{1}
\setcounter{figure}{0}
\setcounter{section}{0}
\setcounter{table}{0}

\renewcommand{\thesection}{\Alph{section}} % 将章节编号设置为字母A, B, C等
\renewcommand{\thesubsection}{\Alph{section}\arabic{subsection}} % 将小节编号设置为A1, A2, A3等

% Redefine figure numbering
\renewcommand{\thefigure}{S\arabic{figure}}
\renewcommand{\thetable}{S\arabic{table}}

\maketitlesupplementary

\section{Detailed Results of LLM-Free Evaluation}
In this section, we report the detailed results of MLLMs on our proposed LLM-free evaluation framework in~\cref{tab:llm-free-detail}.
In addition to the overall scores reported in the main paper, we also include specific results in three aspects: camera motion, subject action list, and temporal action list. 
The camera motion metric assesses the model's ability to accurately identify and describe viewpoint changes and camera movements.
The subject action list metric focuses on the actions of individual subjects, while the temporal action list evaluates the understanding and description of all subjects in terms of chronological order. 
For the subject and temporal action lists, we further consider the action matching degree and action sequence separately.
Specifically, action matching measures the accuracy and detailedness of models' descriptions of the actions each subject performs. Action sequence checks whether these actions are reported in the correct temporal order. 

As we can conclude from~\cref{tab:llm-free-detail}, while most models achieve high scores (over 85\%) in both subject-specific and temporal action sequences, they have significant difficulty with action matching (typically scoring below 40\%). 
This suggests that current MLLMs can reasonably determine the order of actions when they identify them correctly, but they fail to comprehensively recognize part of motions happening in videos.
Besides, our fine-tuned Qwen2.5-VL-7B+FAVOR-Train model shows consistent improvements across all metrics compared to the baseline, with particularly notable performance gains in camera motion understanding (+16.54\%) and action match capabilities (+6.49\% for subject actions and +5.81\% for temporal actions). 
These improvements further demonstrate the effectiveness of our FAVOR-Train dataset in enhancing fine-grained motion understanding.

\renewcommand{\arraystretch}{1.2}
\begin{table*}[t]
\small
\resizebox{\linewidth}{!}{
\begin{tabular}{lcccccccc}
\toprule
%\specialrule{0em}{0.3pt}{0.3pt}
\multirow{2}{*}{\textbf{Methods}} & \multirow{2}{*}{\textbf{Date}} & \multirow{2}{*}{\textbf{Input}} & \multirow{2}{*}{\textbf{Camera Motion}} & \multicolumn{2}{c}{\textbf{Subject Action List}} & \multicolumn{2}{c}{\textbf{Temporal Action List}} & \multirow{2}{*}{\textbf{Overall}} \\ 
\specialrule{0em}{0.3pt}{0.3pt}
\cmidrule(r){5-6} \cmidrule(r){7-8}
\specialrule{0em}{0.1pt}{0.1pt}
&~&~& ~ & Action Match & Action Sequence & Action Match & Action Sequence & ~ \\
\specialrule{0em}{0.3pt}{0.3pt}
\hline
\rowcolor[HTML]{eff0f1}Full mark &-- &-- &100 &100 &100 &100 &100 &100 \\
%\rowcolor[HTML]{eff0f1}Random &-- &-- &20 &20 &20 &20 &20 &20 \\

\hline
 \rowcolor[HTML]{E3F8F8}\multicolumn{9}{l}{\textcolor{gray}{{\textit{\textbf{Proprietary MLLMs}}}}}\\
\rowcolor[HTML]{E3F8F8}Gemini-1.5-Pro~\cite{team2024gemini} & 2024-04 & 1 fps$^*$ & 50.11 & 40.73 & 93.55 & 42.39 & 93.77 & 52.91 \\
\rowcolor[HTML]{E3F8F8}GPT-4o~\cite{hurst2024gpt4o} & 2024-08 & 1 fps$^*$ & 51.83 & 37.08 & 93.62 & 39.52 & 94.67 & 49.50 \\
\rowcolor[HTML]{E3F8F8}Claude-3.7-Sonnet~\cite{Claude3.7} & 2025-02 & 1 fps$^*$ & 51.62 & 26.65 & 90.31 & 29.98 & 91.11 & 43.03 \\
          
\hline
        
 \rowcolor[HTML]{FFF5F5}\multicolumn{9}{l}{\textcolor{gray}{{\textit{\textbf{Open-source MLLMs}}}}} \\

\rowcolor[HTML]{FFF5F5}Video-LLaVA-7B~\cite{lin2024video-llava} & 2023-11 & 8 frms & 49.73 & 25.15 & 89.85 & 25.44 & 89.07 & 41.36 \\
\rowcolor[HTML]{FFF5F5}LLaVA-NeXT-Video-7B~\cite{zhang2024llavanextvideo} & 2024-05 & 8 frms & 46.20 & 14.11 & 84.40 & 13.50 & 66.75 & 29.48 \\
\rowcolor[HTML]{FFF5F5}LLaVA-NeXT-Video-34B~\cite{zhang2024llavanextvideo} & 2024-05 & 8 frms & 50.99 & 23.87 & 89.50 & 23.57 & 86.11 & 39.41 \\
\rowcolor[HTML]{FFF5F5}Tarsier-7B~\cite{wang2024tarsier} & 2024-07 & 8 frms & 62.45 & 29.54 & 90.05 & 28.56 & 88.33 & 46.25 \\
\rowcolor[HTML]{FFF5F5}Tarsier-34B~\cite{wang2024tarsier} & 2024-07 & 8 frms & 63.28 & 30.67 & 91.06 & 29.57 & 89.14 & 47.13 \\
\rowcolor[HTML]{FFF5F5}Aria~\cite{aria} & 2024-10 & 8 frms & 44.79 & 27.61 & 90.92 & 28.87 & 90.22 & 42.78 \\
\rowcolor[HTML]{FFF5F5}InternVL2.5-2B~\cite{chen2024internvl2.5} & 2024-12 & 8 frms & 50.18 & 27.98 & 91.61 & 29.21 & 89.67 & 43.23 \\
\rowcolor[HTML]{FFF5F5}InternVL2.5-8B~\cite{chen2024internvl2.5} & 2024-12 & 8 frms & 52.84 & 28.79 & 91.24 & 30.49 & 90.80 & 44.18 \\
\rowcolor[HTML]{FFF5F5}InternVL2.5-78B~\cite{chen2024internvl2.5} & 2024-12 & 8 frms & 56.79 & 28.19 & 91.19 & 30.18 & 91.11 & 44.01 \\
\rowcolor[HTML]{FFF5F5}Tarsier2-Recap-7B~\cite{yuan2025tarsier2} & 2024-12 & 16 frms & 63.41 & 43.41 & 93.66 & 46.54 & 95.43 & 56.58 \\
\rowcolor[HTML]{FFF5F5}LLaVA-Video-7B-Qwen2~\cite{zhang2024llava-video} & 2024-10 & 64 frms & 48.37 & 30.59 & 91.69 & 31.73 & 91.30 & 45.41 \\
\rowcolor[HTML]{FFF5F5}LLaVA-Video-72B-Qwen2~\cite{zhang2024llava-video} & 2024-10 & 64 frms & 47.78 & 31.01 & 91.98 & 33.43 & 92.33 & 46.06 \\

\rowcolor[HTML]{FFF5F5}VideoChat-Flash-Qwen2-7B~\cite{li2024videochatflash} & 2025-01 & 1 fps & 66.00 & 24.00 & 87.39 & 22.83 & 77.75 & 40.82 \\
\rowcolor[HTML]{FFF5F5}VideoLLaMA3-2B~\cite{zhang2025videollama3} & 2025-01 & 1 fps & 50.90 & 24.49 & 88.93 & 24.79 & 78.06 & 39.29 \\
\rowcolor[HTML]{FFF5F5}VideoLLaMA3-7B~\cite{zhang2025videollama3} & 2025-01 & 1 fps & 53.19 & 33.04 & 92.44 & 34.33 & 92.52 & 48.63 \\
\rowcolor[HTML]{FFF5F5}Qwen2.5-VL-3B~\cite{yang2024qwen2.5} & 2025-01 & 1 fps & 56.64 & 31.51 & 91.88 & 33.05 & 92.04 & 47.32 \\
\rowcolor[HTML]{FFF5F5}Qwen2.5-VL-7B~\cite{yang2024qwen2.5} & 2025-01 & 1 fps & 50.04 & 36.18 & 90.17 & 38.68 & 90.66 & 48.46 \\
\rowcolor[HTML]{FFF5F5}Qwen2.5-VL-72B~\cite{yang2024qwen2.5} & 2025-01 & 1 fps & 53.92 & 36.75 & 93.32 & 39.60 & 94.53 & 49.72 \\

\hline

\rowcolor[HTML]{F1F6EC}Qwen2.5-VL-7B+FAVOR-Train & -- & 1 fps & 66.58 & 42.67 & 91.55 & 44.49 & 92.32 & 56.33 \\

\bottomrule
\end{tabular}
}
\vspace{-0.2cm}
\caption{Detailed LLM-free evaluation results of MLLMs on FAVOR-Bench. The evaluation is divided into three main categories: Camera Motion, Subject Action List (including Subject Action Match and Subject Action Sequence), and Temporal Action List (including Temporal Action Match and Temporal Action Sequence). Due to API response limitations, the video input of proprietary MLLMs are restricted to 16 frames if the video is longer than 16 seconds (denoted as ``1 fps$^*$").
During the calculation of the overall scores, subject and temporal action matching are assigned the highest weights.}

\label{tab:llm-free-detail}
\end{table*}

\section{More Details of FAVOR-Bench}
% 放prompt，标注例子啥的，标注例子放在哪里？
\subsection{More Data Statistics}
\begin{figure*}
    \centering
    \includegraphics[width=\linewidth]{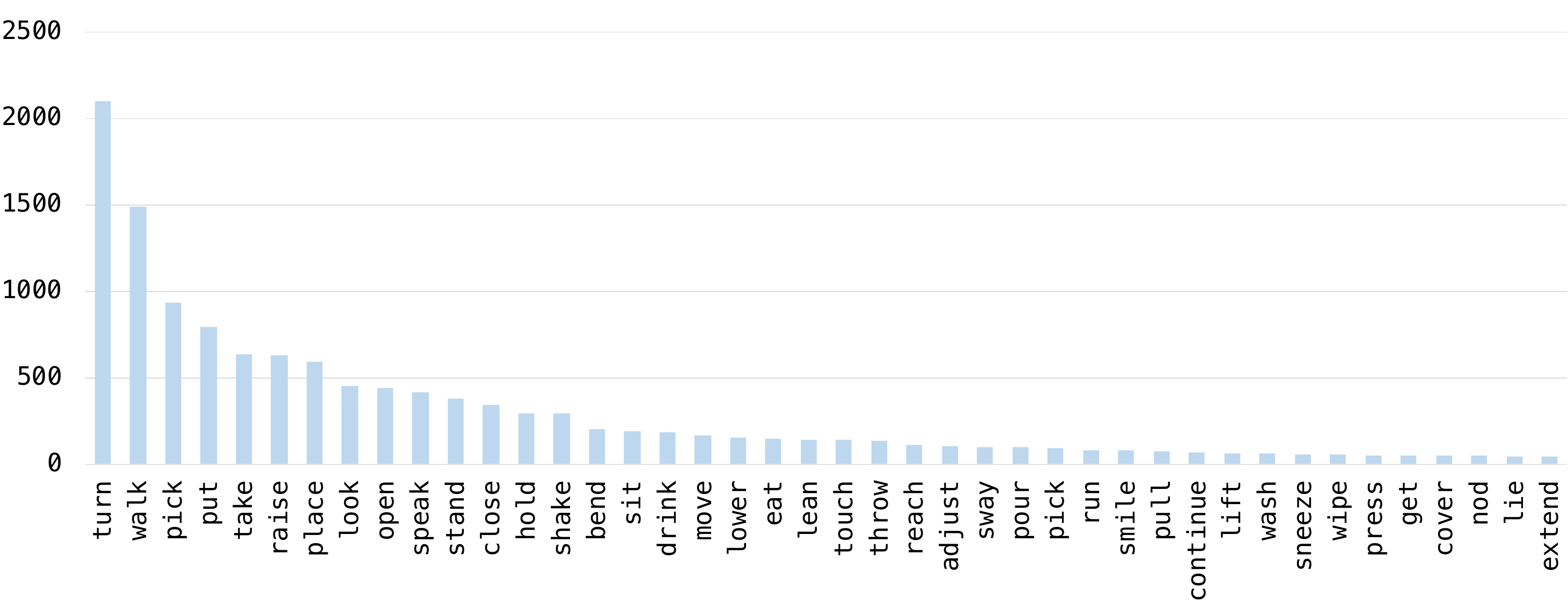}
    \caption{Statistics of motion words with the highest frequency in FAVOR-Bench.}
    \label{appendix_fig:long tail}
\end{figure*}

% 选项分布、动词分布、词云
In this subsection, we provide more data statistics of FAVOR-Bench.
In~\cref{appendix_fig:more statistics}, we provide the distribution of video durations and the number of questions for each video, as well as the index distribution of correct answers.
\Cref{appendix_fig:long tail} exhibits the statistics of motion words with the highest frequency in FAVOR-Bench.

\begin{figure*}
    \centering
    \includegraphics[width=\linewidth]{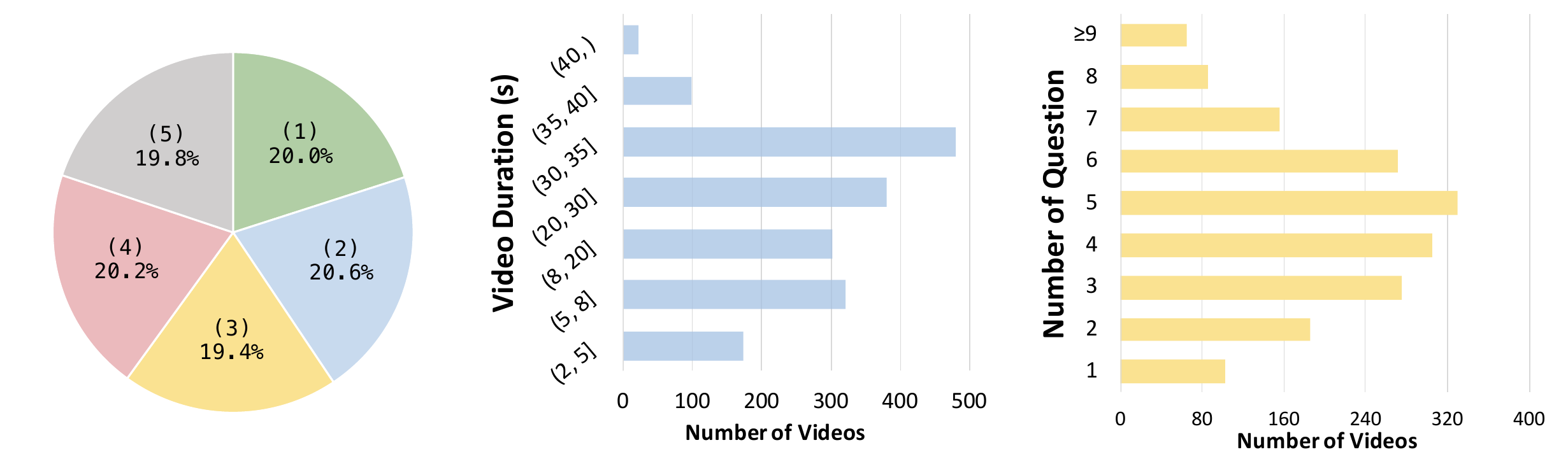}
    \caption{More data statistics of FAVOR-Bench. \textbf{Left:} Index distribution of correct answers for the close-ended tasks. For example, ``(1)" indicates that the correct option is ranked first.
    \textbf{Middle:} Video duration distribution of FAVOR-Bench.
    \textbf{Right:} Question number distribution for videos of FAVOR-Bench.}
    \label{appendix_fig:more statistics}
\end{figure*}

% \begin{figure*}
%     \centering
%     \includegraphics[width=\linewidth]{figures/FAVOR/favor bench.pdf}
%     \caption{}
%     \label{appendix_fig:favor bench}
% \end{figure*}

\subsection{Process of Data Curation}
% 数据收集、标注规则、结构化标注示例
The detailed data curation process for different video sources are as follows. The purpose of this process is getting videos with high quality and dynamic motions.
\begin{itemize}
    \item \textbf{Daily-life record}: For this subset, we sample videos from Charades~\cite{sigurdsson2016hollywood_charades}. This dataset is publicly available and contains various types of human actions and interactions in daily life. 
    Video lengths, number of subjects and density of motions, and interactions of this dataset are relatively moderate. 868 videos are sampled from the subset with the highest quality score (quality scores are offered by Charades). 
    \item \textbf{TV-series \& animation}: For more diverse distributions of characters, motions and scenarios, we collect a corpus of data from TV series and animations to form new subsets. 
    Raw videos with short side resolution greater than 480 are first cropped into segments of a few to tens of seconds. In order to avoid scene transition in clips, adaptive detector algorithm from scene detector library is adopted to generate list of scenes before generating the clip start and end times. Then optical flow calculation with OpenCV is utilized to filter out segments with little motion. 
    Additionally, we manually curate the segments, selecting approximately one out of every twenty clips from a large pool of candidates. 
    Our selection criteria focus on avoiding clips with overly simplistic motions (for example, a person talking throughout the whole video clip) or an excessive number of dynamic subjects in the frame to ensure high-quality fine-grained annotations. 
    Specifically, we exclude segments that only featured simple motions, such as talking heads or basic walking movements. 
    % Besides, we aimed for a balanced representation of single-subject and multi-subject scenarios. 
    In addition, the video selection process considers the balanced distribution of single-subject and multi-subject scenarios. 
    Finally, 574 clips from TV-series and 138 clips from animations are selected.
    \item \textbf{Egocentric}: In order to further expand the distribution of motions and present new challenges, a certain amount of egocentric videos have been included in our benchmark. 
    These videos capture unique interaction patterns and contextual information, occupying a certain proportion in real-world scenarios. 
    We select EgoTaskQA~\cite{jia2022egotaskqa} as the data source and randomly select 196 egocentric videos featuring kitchen and daily activities.
\end{itemize}

\begin{figure*}
    \centering
    \includegraphics[width=0.9\linewidth]{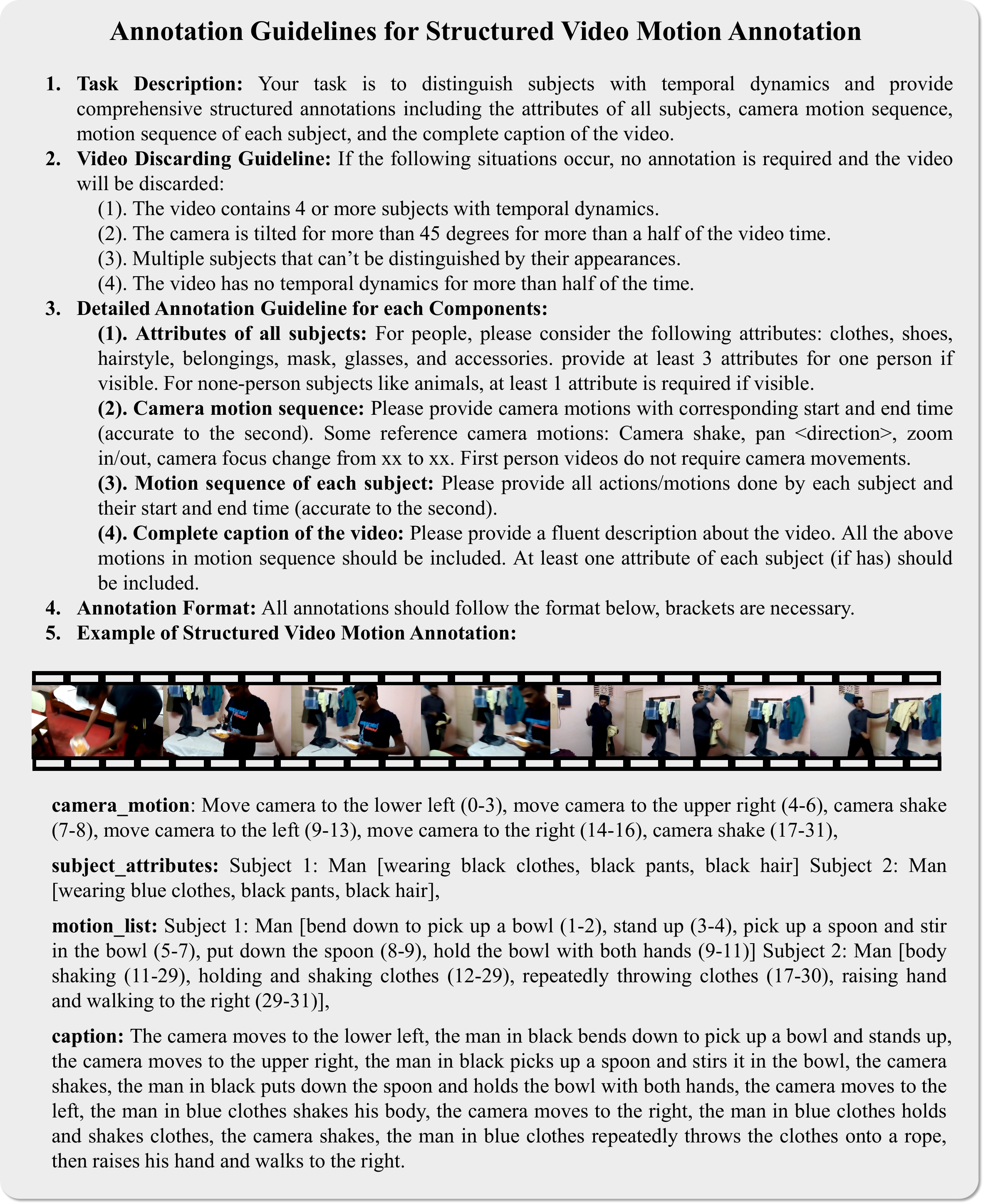}
    \caption{Annotation guidelines of structured video motion annotation and a case of our testing video and corresponding annotations.}
    \label{fig:supp_guideline}
\end{figure*}

To get high quality structured manual annotation, we build a comprehensive annotation guideline as shown in~\cref{fig:supp_guideline}.

\subsection{Prompt for Automatic QA Generation}
\label{app:qa generation}
This subsection provides the prompt templates for automatically generating QA pairs. 
These templates are carefully designed to ensure high-quality questions that thoroughly evaluate the fine-grained video motion understanding capabilities of MLLMs.
For each of the six close-ended tasks in FAVOR-Bench, we use distinct prompt to guide DeepSeek-R1~\cite{guo2025deepseek} in generating diverse and challenging multiple-choice questions. 
Each template is adjusted according to the type of the task and emphasizes different aspects of fine-grained motion understanding.
Tables \ref{tab:as_prompt}-\ref{tab:sad_prompt} present the detailed prompt templates for each task type.

\begin{table*}[ht]
\centering
\begin{tabular}{|p{0.97\linewidth}|}
\hline
\textbf{Prompt Template: Generating QA Pairs for Action Sequence (AS) Task} \\
\hline
You are a professional question designer focusing on temporal dynamics in videos, including camera movements, motions, activities, and interactions, rather than static content. You will receive detailed annotations about the temporal details of the entire video, with duration markers in parentheses after ``camera\_motion" and ``motion\_list". Based on these annotations, design 3 multiple-choice questions around the ``Action Sequence" theme to test models' fine-grained video motion understanding, particularly:
\begin{itemize}
    \item Understanding and analysis of temporal logic, requiring precise identification of action sequences and comprehension of dynamic continuity.
\end{itemize}

\vspace{3mm}
Additionally, follow these question design guidelines:
\begin{enumerate}
    \item Multiple-choice questions should include 5 options (without A,B,C option labels), ensuring exactly one correct answer. If descriptions are ambiguous (e.g., timeline or content), prioritize answer uniqueness.
    \item Focus on representative and significant events or motions, avoiding excessive details or traps.
    \item Distractor generation standards:
    \begin{itemize}
        \item Avoid excessive similarity between options.
        \item Exclude options requiring subjective inference (emotions, intentions, etc.).
        \item Distractors can be outside the annotated motion list. Design relevant distractors, but ensure they don't affect the uniqueness of the correct answer.
    \end{itemize}
    \item Subject identification standards:
    \begin{itemize}
        \item When involving multiple subjects, use subject attributes (e.g., ``man wearing black clothes with gold patterns") for reference.
        \item Avoid abstract identifiers like ``subject 1" or ``person A".
        \item Extract key visual features (color, position, clothing, etc.) from the original annotations.
    \end{itemize}
    \item Avoid questions about specific moments, such as ``When...?" or ``At which second...?"
    \item Different questions should cover more video content, avoiding repetitive questioning.
    \item Avoid questions that can be answered correctly just from the question wording, or from a single frame, without watching the entire video.
    \item Include 1 multiple-choice question asking about a subject's complete behavioral sequence in the video, creating distractors by rearranging the sequence. Connect different behaviors/actions with arrows.
\end{enumerate} \\
\hline
\end{tabular}
\caption{Prompt template for Action Sequence (AS) task.}
\label{tab:as_prompt}
\end{table*}

\begin{table*}[ht]
\centering
\begin{tabular}{|p{0.97\linewidth}|}
\hline
\textbf{Prompt Template: Generating QA Pairs for Camera Motion (CM) Task} \\
\hline
You are a professional question designer focusing on temporal dynamics in videos, including camera movements, motions, activities, and interactions, rather than static content. You will receive detailed annotations about the temporal details of the entire video, with duration markers in parentheses after ``camera\_motion" and ``motion\_list". Based on these annotations, design 3 multiple-choice questions around the ``Camera Motion" theme to test models' fine-grained video motion understanding, particularly:
\begin{itemize}
    \item Understanding camera movement direction and focus changes in the video.
\end{itemize}

\vspace{3mm}
Additionally, follow these question design guidelines:
\begin{enumerate}
    \item If a video's ``camera\_motion" has only one element, such as ``camera\_motion": ``static", or ``camera\_motion": ``camera shaking (0-22)", skip this video and don't generate any content.
    \item Multiple-choice questions should include 5 options (without A,B,C option labels), ensuring exactly one correct answer. If descriptions are ambiguous, prioritize answer uniqueness.
    \item Focus only on camera movements and focus changes. Concentrate on representative and significant events or motions, avoiding excessive details or traps.
    \item Distractor generation standards:
    \begin{itemize}
        \item Avoid excessive similarity between options.
        \item Exclude options requiring subjective inference (emotions, intentions, etc.).
        \item Distractors can be outside the annotated motion list. Design relevant distractors, but ensure they don't affect the uniqueness of the correct answer.
    \end{itemize}
    \item Subject identification standards:
    \begin{itemize}
        \item When involving multiple subjects, use subject attributes (e.g., ``man wearing black clothes with gold patterns") for reference.
        \item Avoid abstract identifiers like ``subject 1" or ``person A".
        \item Extract key visual features (color, position, clothing, etc.) from the original annotations.
    \end{itemize}
    \item Avoid questions about specific moments, such as ``When...?" or ``At which second...?"
    \item Different questions should cover more video content, avoiding repetitive questioning.
    \item Questions and answers should not include any specific time information, such as ``during the camera shake phase (0-3 seconds)"
    \item If ``camera\_motion" doesn't mention focus, designed questions should not include focus.
    \item Avoid questions that can be answered correctly just from the question wording, or from a single frame, without watching the entire video.
\end{enumerate} \\
\hline
\end{tabular}
\caption{Prompt template for Camera Motion (CM) task.}
\label{tab:cm_prompt}
\end{table*}

\begin{table*}[ht]
\centering
\begin{tabular}{|p{0.97\linewidth}|}
\hline
\textbf{Prompt Template: Generating QA Pairs for Holistic Action Classification (HAC) Task} \\
\hline
You are a professional question designer focusing on temporal dynamics in videos, including camera movements, motions, activities, and interactions, rather than static content. You will receive detailed annotations about the temporal details of the entire video, with duration markers in parentheses after ``camera\_motion" and ``motion\_list". Based on these annotations, design 3 multiple-choice questions around the ``Holistic Action Classification" theme to test models' fine-grained video motion understanding, particularly:
\begin{itemize}
    \item Emphasizing overall summarization ability, requiring distillation of core behaviors from the entire video.
\end{itemize}

\vspace{3mm}
Additionally, follow these question design guidelines:
\begin{enumerate}
    \item Multiple-choice questions should include 5 options (without A,B,C option labels), ensuring exactly one correct answer. If descriptions are ambiguous, prioritize answer uniqueness.
    \item Focus on the main behaviors and motions throughout the entire video, which should cover most of the video duration or occupy more than half of the caption length. Don't ask about motions occurring only in local time segments of the video, avoiding expressions like ``in the first half of the video" or ``at the end of the video."
    \item Distractor generation standards:
    \begin{itemize}
        \item Avoid excessive similarity between options (e.g., ``chin moving up and down" vs. ``slight opening and closing").
        \item Exclude options requiring subjective inference (emotions, intentions, etc.).
        \item Distractors can be outside the annotated motion list. Design relevant distractors, but ensure they don't affect the uniqueness of the correct answer.
    \end{itemize}
    \item Subject identification standards:
    \begin{itemize}
        \item When involving multiple subjects, use subject attributes (e.g., ``man wearing black clothes with gold patterns") for reference.
        \item Avoid abstract identifiers like ``subject 1" or ``person A".
        \item Extract key visual features (color, position, clothing, etc.) from the original annotations.
    \end{itemize}
    \item Avoid questions about specific moments, such as ``When...?" or ``At which second...?"
    \item Different questions should cover more video content, avoiding repetitive questioning.
    \item Avoid questions that can be answered correctly just from the question wording, or from a single frame, without watching the entire video.
\end{enumerate} \\
\hline
\end{tabular}
\caption{Prompt template for Holistic Action Classification (HAC) task.}
\label{tab:hac_prompt}
\end{table*}

\begin{table*}[ht]
\centering
\begin{tabular}{|p{0.97\linewidth}|}
\hline
\textbf{Prompt Template: Generating QA Pairs for Multiple Action Detail (MAD) Task} \\
\hline
You are a professional question designer focusing on temporal dynamics in videos, including camera movements, motions, activities, and interactions, rather than static content. You will receive detailed annotations about the temporal details of the entire video, with duration markers in parentheses after ``camera\_motion" and ``motion\_list". Based on these annotations, design 3 multiple-choice questions around the ``Multiple Action Detail" theme to test models' fine-grained video motion understanding, particularly:
\begin{itemize}
    \item Multi-moment information comparison and analysis ability, requiring understanding of object state changes at different times, subject interactions with multiple objects at different times, or subject multiple interactions with the same object, emphasizing contrast and associative reasoning.
\end{itemize}

\vspace{3mm}
Additionally, follow these question design guidelines:
\begin{enumerate}
    \item Multiple-choice questions should include 5 options (without A,B,C option labels), ensuring exactly one correct answer. If descriptions are ambiguous, prioritize answer uniqueness.
    \item Focus on representative and significant events or motions, avoiding excessive details or traps.
    \item Distractor generation standards:
    \begin{itemize}
        \item Avoid excessive similarity between options (e.g., ``chin moving up and down" vs. ``slight opening and closing").
        \item Exclude options requiring subjective inference (emotions, intentions, etc.).
        \item Distractors can be outside the annotated motion list. Design relevant distractors, but ensure they don't affect the uniqueness of the correct answer.
    \end{itemize}
    \item Subject identification standards:
    \begin{itemize}
        \item When involving multiple subjects, use subject attributes (e.g., ``man wearing black clothes with gold patterns") for reference.
        \item Avoid abstract identifiers like ``subject 1" or ``person A".
        \item Extract key visual features (color, position, clothing, etc.) from the original annotations.
    \end{itemize}
    \item Avoid questions about specific moments, such as ``When...?" or ``At which second...?"
    \item Different questions should cover more video content, avoiding repetitive questioning.
    \item If a subject interacted with multiple objects, include 1 multiple-choice question asking about which items a subject interacted with at different times; otherwise, this is not needed.
    \item Avoid questions that can be answered correctly just from the question wording, or from a single frame, without watching the entire video.
\end{enumerate} \\
\hline
\end{tabular}
\caption{Prompt template for Multiple Action Detail (MAD) task.}
\label{tab:mad_prompt}
\end{table*}

\begin{table*}[ht]
\centering
\begin{tabular}{|p{0.97\linewidth}|}
\hline
\textbf{Prompt Template: Generating QA Pairs for Single Action Detail (SAD) Task} \\
\hline
You are a professional question designer focusing on temporal dynamics in videos, including camera movements, motions, activities, and interactions, rather than static content. You will receive detailed annotations about the temporal details of the entire video, with duration markers in parentheses after ``camera\_motion" and ``motion\_list". Based on these annotations, design 3 multiple-choice questions around the ``Single Action Detail" theme to test models' fine-grained video motion understanding, particularly:
\begin{itemize}
    \item Testing detail capturing ability, understanding subject interaction with a specific object, or subject/object state at a specific moment.
\end{itemize}

\vspace{3mm}
Additionally, follow these question design guidelines:
\begin{enumerate}
    \item Multiple-choice questions should include 5 options (without A,B,C option labels), ensuring exactly one correct answer. If descriptions are ambiguous, prioritize answer uniqueness.
    \item Focus on representative and significant events or motions, avoiding excessive details or traps.
    \item Distractor generation standards:
    \begin{itemize}
        \item Avoid excessive similarity between options (e.g., ``chin moving up and down" vs. ``slight opening and closing").
        \item Exclude options requiring subjective inference (emotions, intentions, etc.).
        \item Distractors can be outside the annotated motion list. Design relevant distractors, but ensure they don't affect the uniqueness of the correct answer.
    \end{itemize}
    \item Subject identification standards:
    \begin{itemize}
        \item When involving multiple subjects, use subject attributes (e.g., ``man wearing black clothes with gold patterns") for reference.
        \item Avoid abstract identifiers like ``subject 1" or ``person A".
        \item Extract key visual features (color, position, clothing, etc.) from the original annotations.
    \end{itemize}
    \item Avoid questions about specific moments, such as ``When...?" or ``At which second...?"
    \item Different questions should cover more video content, avoiding repetitive questioning.
    \item Avoid questions that can be answered correctly just from the question wording, or from a single frame, without watching the entire video.
\end{enumerate} \\
\hline
\end{tabular}
\caption{Prompt template for Single Action Detail (SAD) task.}
\label{tab:sad_prompt}
\end{table*}

\begin{table*}[ht]
\centering
\begin{tabular}{|p{0.97\linewidth}|}
\hline
\textbf{Prompt Template: Generating QA Pairs for Non-subject Motion (NSM) Task} \\
\hline
You are a professional question designer focusing on temporal dynamics in videos, including camera movements, motions, activities, and interactions, rather than static content. You will receive detailed annotations about the temporal details of the entire video, with duration markers in parentheses after ``camera\_motion" and ``motion\_list". Based on these annotations, design 3 multiple-choice questions around the ``Non-subject Motion" theme to test models' fine-grained video motion understanding, particularly:
\begin{itemize}
    \item Testing environmental awareness ability, requiring attention to secondary elements (such as background objects, background characters) movement and behavior, as supplementary information for fine-grained video motion understanding.
\end{itemize}

\vspace{3mm}
Additionally, follow these question design guidelines:
\begin{enumerate}
    \item Multiple-choice questions should include 5 options (without A,B,C option labels), ensuring exactly one correct answer. If descriptions are ambiguous, prioritize answer uniqueness.
    \item Focus on representative and significant events or motions, avoiding excessive details or traps.
    \item Distractor generation standards:
    \begin{itemize}
        \item Avoid excessive similarity between options (e.g., ``chin moving up and down" vs. ``slight opening and closing").
        \item Exclude options requiring subjective inference (emotions, intentions, etc.).
        \item Distractors can be outside the annotated motion list. Design relevant distractors, but ensure they don't affect the uniqueness of the correct answer.
    \end{itemize}
    \item Subject identification standards:
    \begin{itemize}
        \item When involving multiple subjects, use subject attributes (e.g., ``man wearing black clothes with gold patterns") for reference.
        \item Avoid abstract identifiers like ``subject 1" or ``person A".
        \item Extract key visual features (color, position, clothing, etc.) from the original annotations.
    \end{itemize}
    \item Avoid questions about specific moments, such as ``When...?" or ``At which second...?"
    \item Different questions should cover more video content, avoiding repetitive questioning.
    \item Avoid questions that can be answered correctly just from the question wording, or from a single frame, without watching the entire video.
\end{enumerate} \\
\hline 
\end{tabular}
\caption{Prompt template for Non-subject Motion (NSM) task.}
\label{tab:nsa_prompt}
\end{table*}

\subsection{Prompt for GPT-Assisted Evaluation}
For the GPT-assisted evaluation, we design a comprehensive prompt to instruct the powerful GPT-4o to directly compare and evaluate model responses from two critical dimensions of fine-grained video motion understanding: correctness and detailedness. 
The correctness dimension assesses whether the model accurately describes the motions, activities, interactions, and camera movements that occur in the video. 
The detailedness dimension evaluates how comprehensively the model captures the temporal dynamics, including the execution manner of actions, camera movements, and interactions. 
To improve the robustness of the evaluation, we develop detailed criteria for different ratings respectively. 
\Cref{tab:gpt_eval_prompt1,tab:gpt_eval_prompt2} present the complete prompt template used in this paper.

\begin{table*}[ht]
\centering
\renewcommand{\arraystretch}{0.9}
\small
\begin{tabular}{|p{0.97\linewidth}|}
\hline
\textbf{Prompt Template for GPT-Assisted Evaluation (1/2)}\\
\hline
\vspace{0.1mm}

Please act as a professional video motion analysis expert to evaluate models' fine-grained motion understanding capabilities in videos. You will compare the model-generated description (Response) with human-annotated standard description (Caption), and rate the model's performance on two dimensions, ``Correctness" and ``Detailedness", each on a scale from 1 to 10.
Remember that the model received this instruction: ``Please analyze and describe the temporal dynamics in this video, focusing on camera movements, motions, activities, and interactions, rather than static content."

\vspace{2mm}
\textbf{Evaluation Dimension 1: Correctness (1-10 points)} \\
Evaluate whether the model's description of motions, activities, interactions, and camera movements that actually appear in the video is accurate.

\vspace{2mm}
\textbf{Correctness Rating Criteria:}
\vspace{2mm}

\textbf{9-10 points (Extremely High Correctness)}\\
- Completely correct description of all core motions, activities, and interactions in the video;\\
- Correctly identified camera movements and changes;\\
- Completely accurate description of motion temporal relationships and directions;\\
- No errors or only negligible minor inaccuracies.

\vspace{2mm}
\textbf{7-8 points (High Correctness)}\\
- Correctly described most core motions, activities, and interactions in the video;\\
- Basically correctly identified the main camera movements;\\
- Generally accurate description of motion temporal relationships and directions;\\
- 1-2 minor errors that don't affect the overall understanding of video dynamic content.

\vspace{2mm}
\textbf{5-6 points (Medium Correctness)}\\
- Correctly described some core motions, activities, and interactions;\\
- Partially identified camera movements;\\
- Some confusion in motion temporal sequence or direction description;\\
- Several obvious errors, but core motion descriptions remain partially correct.

\vspace{2mm}
\textbf{3-4 points (Low Correctness)}\\
- Correctly described only a few motions or activities;\\
- Incorrect or missing description of camera movements;\\
- Numerous errors in motion temporal sequence or direction description;\\
- Multiple obvious errors, significant misunderstanding of video content.

\vspace{2mm}
\textbf{1-2 points (Extremely Low Correctness)}\\
- Almost no correct description of any actual motions or activities;\\
- Severe misunderstanding of video content, numerous errors or fabricated content;\\
- Completely confused motion temporal sequence;\\
- Description almost completely inconsistent with actual video content.

\vspace{2mm}
\textbf{Evaluation Dimension 2: Detailedness (1-10 points)}\\
Evaluate whether the model comprehensively and thoroughly describes the dynamic content in the video, including temporal dynamics, camera movements, motions, activities, and interaction details.

\vspace{2mm}
\textbf{Detailedness Rating Criteria:}
\vspace{2mm}

\textbf{9-10 points (Extremely High Detailedness)}\\
- Comprehensively captured details of all key motions and activities in the video;\\
- Detailed description of how motions are executed (e.g., speed, force, amplitude);\\
- Complete capture of temporal dynamics and motion transitions;\\
- Precise description of various camera movements and changes (e.g., panning, pushing/pulling, rotation);\\
- In-depth analysis of interaction relationships and dynamic changes in the scene.
\vspace{2mm}

\textbf{7-8 points (High Detailedness)}\\
- Captured details of most key motions and activities in the video;\\
- Described the execution manner of most motions;\\
- Good capture of temporal dynamics and main motion transitions;\\
- Described the main camera movements;\\
- Analyzed the main interaction relationships.
\\
\hline
\end{tabular}
\caption{Prompt template for GPT-assisted evaluation of FAVOR-Bench (Part 1).}
\label{tab:gpt_eval_prompt1}
\end{table*}

\begin{table*}[ht]
\centering
\renewcommand{\arraystretch}{0.9}
\small
\begin{tabular}{|p{0.97\linewidth}|}
\hline
\textbf{Prompt Template for GPT-Assisted Evaluation (2/2)} \\
\hline

\vspace{2mm}

\textbf{5-6 points (Medium Detailedness)}\\
- Captured details of some key motions and activities;\\
- Partially described how motions are executed;\\
- Basic capture of temporal dynamics;\\
- Mentioned some camera movements;\\
- Included some interaction relationship descriptions.
\vspace{2mm}

\textbf{3-4 points (Low Detailedness)}\\
- Provided only basic descriptions of motions and activities, lacking details;\\
- Rarely described how motions are executed;\\
- Brief description of temporal dynamics;\\
- Almost no mention of camera movements;\\
- Insufficient description of interaction relationships.
\vspace{2mm}

\textbf{1-2 points (Extremely Low Detailedness)}\\
- Extremely brief, mentioning only the most basic motions;\\
- No description of how motions are executed;\\
- Missing temporal dynamics;\\
- Completely ignored camera movements;\\
- No description of interaction relationships.

\vspace{2mm}
\textbf{Scoring Guiding Principles:}

1.\textbf{Ignore Static Content Assessment:} Scoring should focus on dynamic content; points should not be heavily deducted for missing static scene descriptions.\\
2.\textbf{Tolerate Expression Differences:} Different expressions (e.g., ``moves to the left" vs. ``walks toward the window") should be considered equivalent if they refer to the same motion.\\
3.\textbf{Correctness First:} If a description is seriously incorrect, it should not receive high scores even if detailed.\\
4.\textbf{Distinguish Between Omissions and Errors:} Not mentioning certain content (omission) should not be treated the same as incorrect descriptions; errors should more severely impact correctness scores.\\
5.\textbf{Distinguish Primary from Secondary:} Correct descriptions of core motions/primary activities are more important than minor details.

\vspace{2mm}
\textbf{Output Format:}

\vspace{2mm}
\textbf{Correctness Analysis} \\
\text{[Detailed analysis of how well the model's description matches the actual video content, pointing out correct aspects and errors]}

\vspace{2mm}
\textbf{Detailedness Analysis} \\
\text{[Detailed analysis of the comprehensiveness and richness of details in the model's description, pointing out detailed aspects and} \\
\text{shortcomings]}

\vspace{2mm}
\textbf{Ratings} \\
Correctness Rating: [Integer from 1-10] \\
Detailedness Rating: [Integer from 1-10]

\vspace{2mm}
\textbf{Human-Annotated Standard Description (Caption)} \\
\{caption\}

\vspace{2mm}
\textbf{Model Response} \\
\{response\}
\vspace{1mm}
\\
\hline
\end{tabular}
\caption{Prompt template for GPT-assisted evaluation of FAVOR-Bench (Part 2).}
\label{tab:gpt_eval_prompt2}
\end{table*}

\section{More Details of FAVOR-Train}
% 训练集和训练的更多信息
%\subsection{Video Collection for FAVOR-Train.}
For third-person perspective videos, all videos are sampled from Koala-36M~\cite{wang2024koala_dataset}. Before sampling, an automatic filtering is performed using the clarity score, aestheric score, motions score and VTSS provided by Koala-36M. We use three different sampling strategies to create three subsets from the filtered 7M videos, which collectively make up the third-person component of our training set:

\begin{itemize}
    \item \textbf{Random sampling}: Simple random sampling is performed to get the first one third of videos.
    \item \textbf{Videos with long-tail actions}: Before sampling of this step, we use Qwen2-72B to extract motion phrases in the caption from Koala-36M of each video. We then traverse the filtered subset video list and maintain a set of motions (each from a video) and keep their distance in the semantic embedding space as far as possible. Thus obtaining a set including long-tail motions
    \item \textbf{Videos with typical motions}: We select nearly 1600 motions from the motion set of previous datasets for caption recognition such as Kinetics~\cite{kay2017kinetics}, ActivityNet~\cite{caba2015activitynet}, and MIT~\cite{monfort2019moments}. Videos whose caption contains these motions are accumulated to form another one third of videos.
\end{itemize}

For first-person perspective videos, videos from EgoTaskQA and Charades-Ego can be directly sampled as training data. However, videos from Egoexo4D and EgoexoLearn are much longer. We adopt a cropping strategy based on their temporal annotations and ensure that all cropped video clips are shorter than 40 seconds.

\section{More Samples of FAVOR-Bench}
\label{sec:more results}

\begin{figure*}
    \centering
    \includegraphics[width=0.9\linewidth]{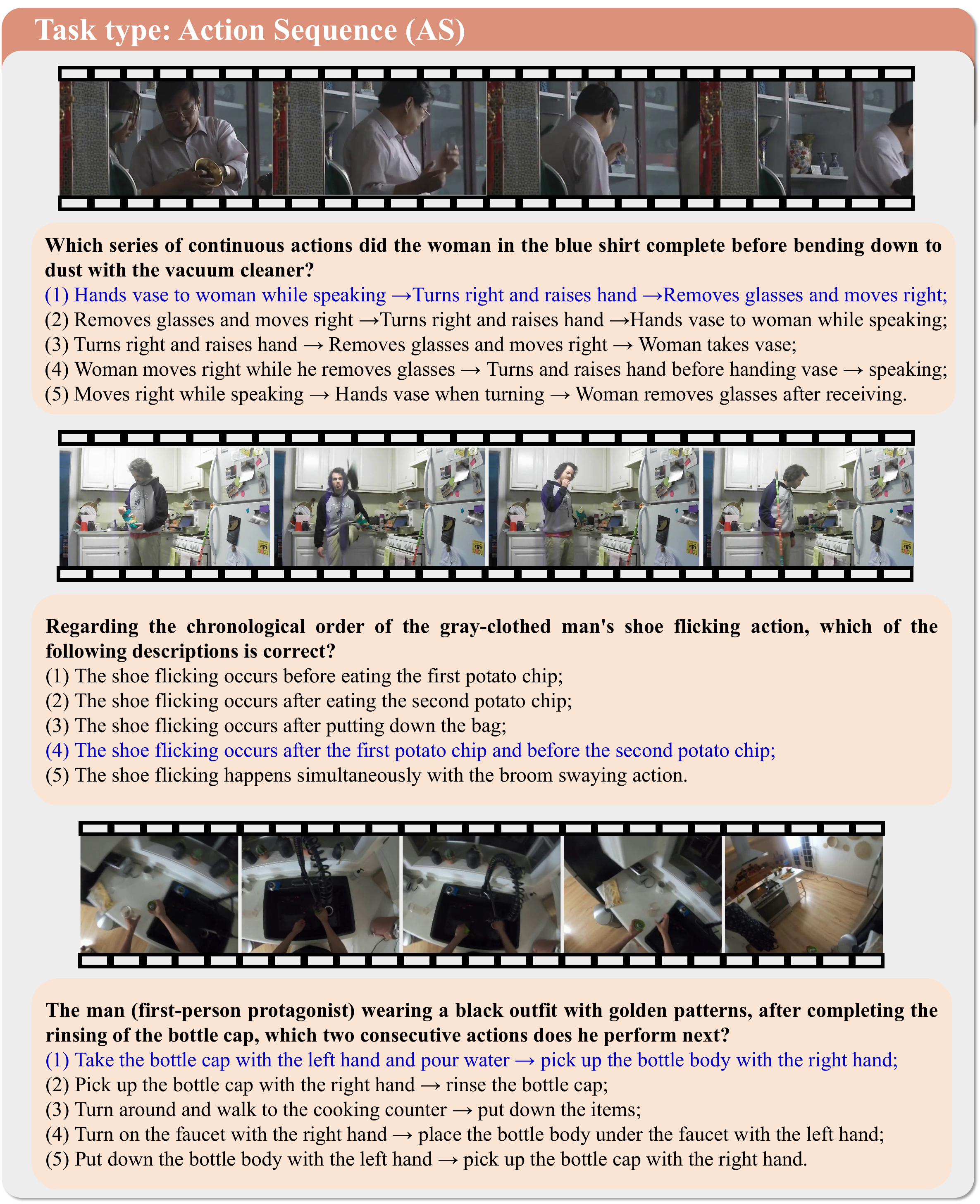}
    \caption{Examples of Action Sequence (AS) task in the close-ended evaluation.}
    \label{fig:supp_as}
\end{figure*}

\begin{figure*}
    \centering
    \includegraphics[width=0.9\linewidth]{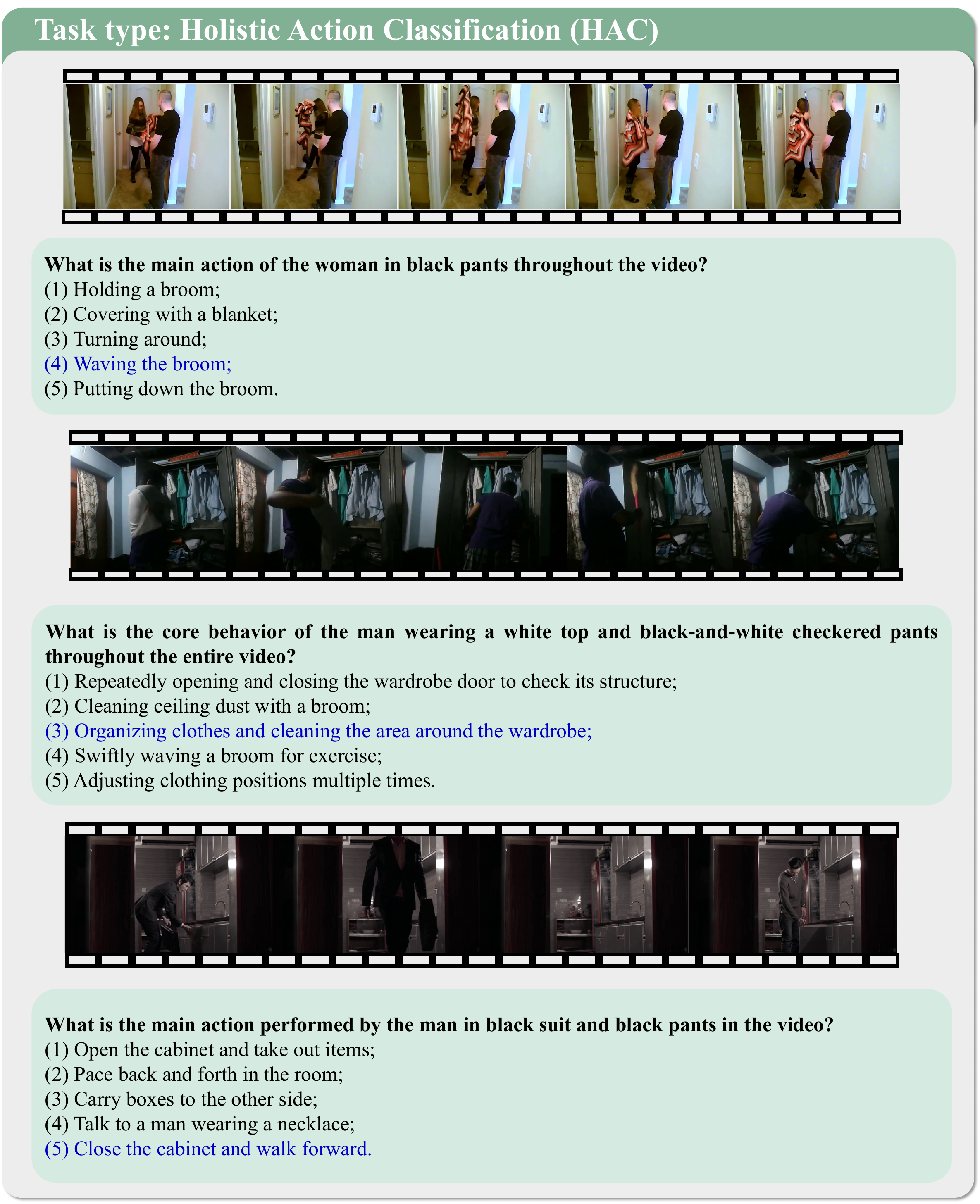}
    \caption{Examples of Holistic Action Classification (HAC) task in the close-ended evaluation.}
    \label{fig:supp_hac}
\end{figure*}

\begin{figure*}
    \centering
    \includegraphics[width=0.9\linewidth]{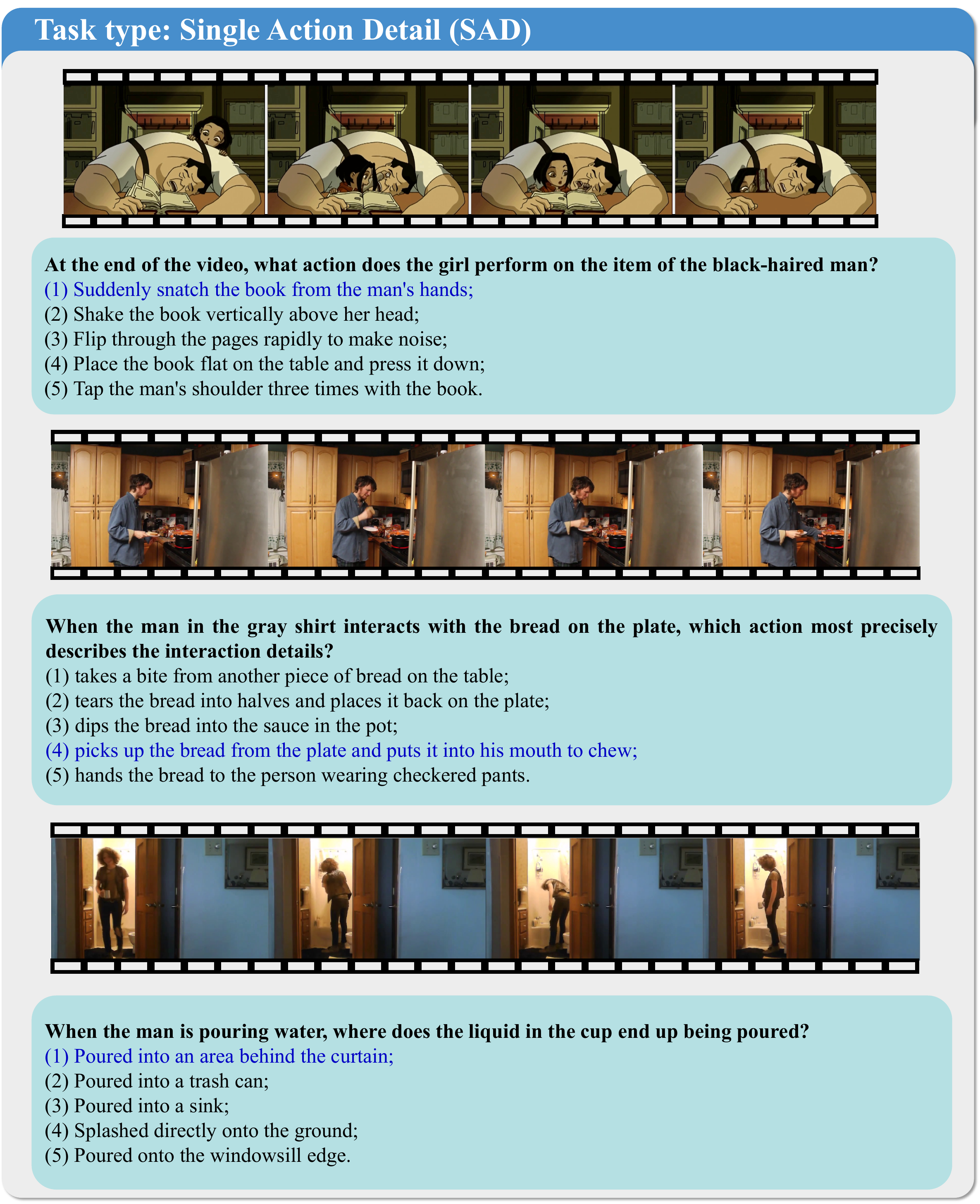}
    \caption{Examples of Single Action Detail (SAD) task in the close-ended evaluation.}
    \label{fig:supp_sad}
\end{figure*}

\begin{figure*}
    \centering
    \includegraphics[width=0.9\linewidth]{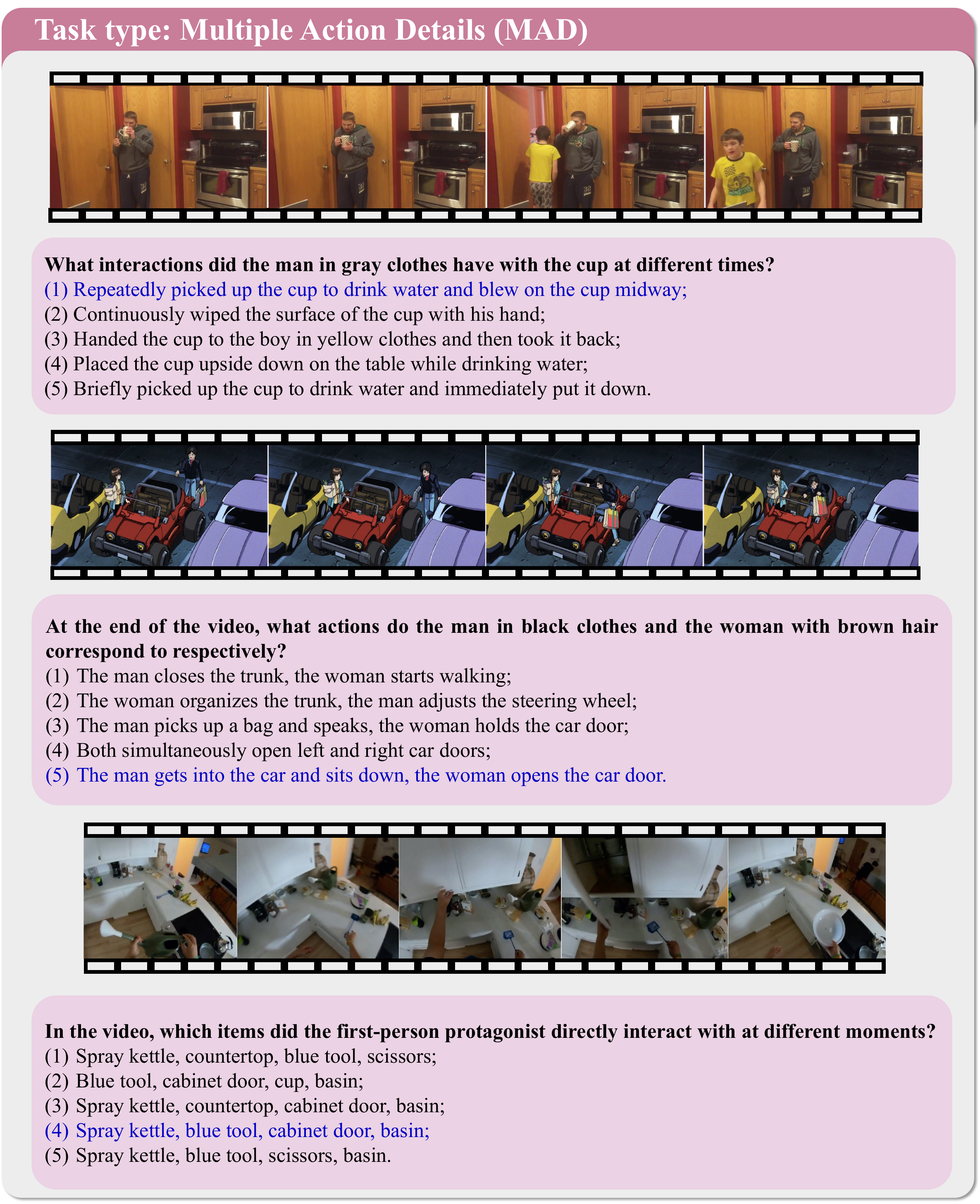}
    \caption{Examples of Multiple Action Details (MAD) task in the close-ended evaluation.}
    \label{fig:supp_mad}
\end{figure*}

\begin{figure*}
    \centering
    \includegraphics[width=0.9\linewidth]{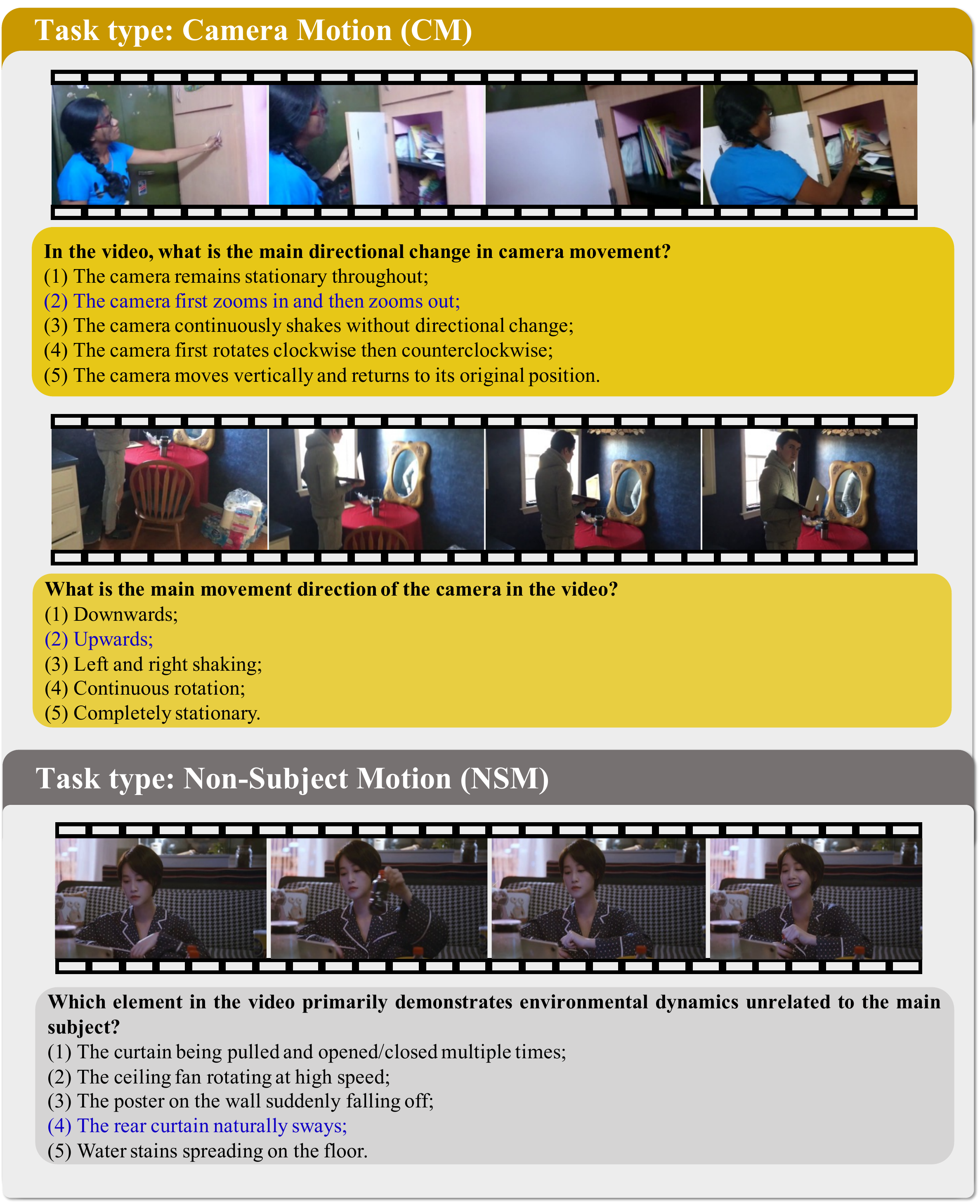}
    \caption{Examples of Camera Motion (CM) and Non-Subject Motion (NSM) tasks in the close-ended evaluation.}
    \label{fig:supp_cm}
\end{figure*}

For better demonstration, we show more samples of videos and their corresponding close-ended questions, options and correct answers for all six tasks in~\cref{fig:supp_as,fig:supp_hac,fig:supp_mad,fig:supp_sad,fig:supp_cm}.

\end{document}